\documentclass[a4paper,fleqn]{cas-sc}

\usepackage[numbers]{natbib}
\usepackage{hyperref}

\usepackage[utf8]{inputenc} 
\usepackage[T1]{fontenc}    
\usepackage{hyperref}       
\usepackage{url}            
\usepackage{booktabs}       
\usepackage{amsfonts}       
\usepackage{nicefrac}       
\usepackage{microtype}      
\usepackage{xcolor}         

\usepackage{times}
\usepackage{epsfig}
\usepackage{graphicx}
\usepackage{amsmath}
\usepackage{amssymb}
\usepackage{color}
\usepackage{algorithm}
\usepackage{algorithmic}
\usepackage{multirow}
\usepackage{pifont}

\newcommand\ie{\emph{i.e.}}
\newcommand\eg{\emph{e.g.}}

\usepackage{colortbl}
\usepackage{xcolor}
\usepackage{graphicx}
\usepackage{subcaption}
\usepackage{multirow}
\usepackage{multicol}
\usepackage{pifont}
\usepackage{makecell}
\usepackage{graphicx}
\usepackage{CJK}
\usepackage[utf8]{inputenc}
\usepackage[T1]{fontenc}
\usepackage{amsmath}
\usepackage{amsfonts}
\usepackage{amssymb}
\usepackage[version=4]{mhchem}
\usepackage{stmaryrd}
\usepackage{bbold}
\usepackage{graphicx, amsmath, amssymb, caption, subcaption, multirow, overpic, textpos}

\usepackage{pifont}
\usepackage{listings}
\usepackage{cite} 
\usepackage{soul}
\usepackage{arydshln}

\usepackage{tabulary,overpic,xcolor}
\definecolor{citecolor}{RGB}{34,139,34}

\def\tsc#1{\csdef{#1}{\textsc{\lowercase{#1}}\xspace}}
\tsc{WGM}
\tsc{QE}
\tsc{EP}
\tsc{PMS}
\tsc{BEC}
\tsc{DE}

\begin{document}
\let\WriteBookmarks\relax
\def\floatpagepagefraction{1}
\def\textpagefraction{.001}

\shorttitle{}

\shortauthors{}

\title[mode = title]{SST: Self-training with Self-adaptive Thresholding for Semi-supervised Learning}



\author[1,2,3]{Shuai Zhao}[orcid=0009-0004-6211-2611]
\ead{zhaoshuai@bit.edu.cn}
\affiliation[1]{
    organization={School of Computer Science and Technology, Beijing Institute of Technology},
    city={Beijing},
    country={China}
    }
\affiliation[2]{
    organization={Beijing Engineering Research Center of High Volume Language Information Processing and Cloud Computing Applications},
    city={Beijing},
    country={China}
    }
\affiliation[3]{
    organization={Southeast Academy of Information Technology, Beijing Institute of Technology},
    state={Fujian},
    country={China}
    }

\author[1,2,3]{Heyan Huang}
\ead{hhy63@bit.edu.cn}
\cormark[1]
\cortext[cor1]{Corresponding author.}

\author[4]{Xinge Li}
\ead{li_xinge@iapcm.ac.cn}
\affiliation[4]{
    organization={Institute of Applied Physics and Computational Mathematics},
    city={Beijing},
    country={China}
    }

\author[5]{Xiaokang Chen}
\ead{chenxiaokang381@starsee.cn}
\affiliation[5]{
    organization={Starsee},
    city={Beijing},
    country={China}
    }

\author[5]{Rui Wang}
\ead{wangrui345@starsee.cn}



\begin{abstract}
Neural networks have demonstrated exceptional performance in supervised learning, benefiting from abundant high-quality annotated data. However, obtaining such data in real-world scenarios is costly and labor-intensive. Semi-supervised learning (SSL) offers a solution to this problem by utilizing a small amount of labeled data along with a large volume of unlabeled data. Recent studies, such as Semi-ViT and Noisy Student, which employ consistency regularization or pseudo-labeling, have demonstrated significant achievements. However, they still face challenges, particularly in accurately selecting sufficient high-quality pseudo-labels due to their reliance on fixed thresholds. Recent methods such as FlexMatch and FreeMatch have introduced flexible or self-adaptive thresholding techniques, greatly advancing SSL research. Nonetheless, their process of updating thresholds at each iteration is deemed time-consuming, computationally intensive, and potentially unnecessary. To address these issues, we propose Self-training with Self-adaptive Thresholding (SST), a novel, effective, and efficient SSL framework. SST integrates with both supervised (Super-SST) and semi-supervised (Semi-SST) learning. SST introduces an innovative Self-Adaptive Thresholding (SAT) mechanism that adaptively adjusts class-specific thresholds based on the model's learning progress. SAT ensures the selection of high-quality pseudo-labeled data, mitigating the risks of inaccurate pseudo-labels and confirmation bias (where models reinforce their own mistakes during training). Specifically, SAT prevents the model from prematurely incorporating low-confidence pseudo-labels, reducing error reinforcement and enhancing model performance. Extensive experiments demonstrate that SST achieves state-of-the-art performance with remarkable efficiency, generalization, and scalability across various architectures and datasets. Notably, Semi-SST-ViT-Huge achieves the best results on competitive ImageNet-1K SSL benchmarks (no external data), with 80.7\% / 84.9\% Top-1 accuracy using only 1\% / 10\% labeled data. Compared to the fully-supervised DeiT-III-ViT-Huge, which achieves 84.8\% Top-1 accuracy using 100\% labeled data, our method demonstrates superior performance using only 10\% labeled data. This indicates a tenfold reduction in human annotation costs, significantly narrowing the performance disparity between semi-supervised and fully-supervised methods. These advancements pave the way for further innovations in SSL and practical applications where obtaining labeled data is either challenging or costly.
\end{abstract}


\maketitle

\setcounter{page}{1}

\section{Introduction}
\label{sec:introduction}

Neural networks have demonstrated exceptional performance in various supervised learning tasks \citep{russakovsky2015imagenet, deng2009imagenet, lin2014microsoft}, benefiting from abundant high-quality annotated data. However, obtaining sufficient high-quality annotated data in real-world scenarios can be both costly and labor-intensive. Therefore, learning from a few labeled examples while effectively utilizing abundant unlabeled data has become a pressing challenge. Semi-supervised learning (SSL) \citep{zhu2005semi, zhu2022introduction, sohn2020fixmatch, van2020survey} offers a solution to this problem by utilizing a small amount of labeled data along with a large volume of unlabeled data. 

Consistency regularization, pseudo-labeling, and self-training are prominent SSL methods. Consistency regularization \citep{laine2016temporal, tarvainen2017mean, miyato2018virtual, xie2020unsupervised, bachman2014learning, rasmus2015semi, berthelot2019mixmatch, sajjadi2016regularization} promotes the model to yield consistent predictions for the same unlabeled sample under various perturbations, thereby enhancing model performance. Pseudo-labeling and self-training \citep{scudder1965probability, mclachlan1975iterative, yarowsky1995unsupervised, riloff1996automatically, riloff2003learning, he2011self, lee2013pseudo, shi2018transductive, iscen2019label, arazo2020pseudo, sohn2020simple, xie2020self, xie2020unsupervised, sohn2020fixmatch, pham2021meta, cai2021exponential, liu2021unbiased, zhang2021flexmatch, wei2021crest, pedronette2021rank, alqahtani2023improving} are closely related in SSL. Pseudo-labeling assigns labels to unlabeled data, while self-training iteratively uses these pseudo-labels to enhance model performance.

Recent studies, such as Semi-ViT \citep{cai2022semi} and Noisy Student \citep{xie2020self}, which employ consistency regularization, pseudo-labeling, or self-training, have demonstrated significant achievements. However, they still face challenges, particularly in accurately selecting sufficient high-quality pseudo-labels due to their reliance on fixed thresholds. Recent methods such as FlexMatch \citep{zhang2021flexmatch} and FreeMatch \citep{wang2023freematch} have introduced flexible or self-adaptive thresholding techniques, greatly advancing SSL research. Nonetheless, their process of updating thresholds at each iteration is deemed time-consuming, computationally intensive, and potentially unnecessary. Moreover, both FlexMatch and FreeMatch predict unlabeled data and update thresholds based on models that are still in training, rather than on converged models with high accuracy. During the early stages of training, using lower-performance models to predict unlabeled data inevitably introduces more inaccurate pseudo-labels, leading to a feedback loop of increasing inaccuracies. This amplifies and reinforces the confirmation bias \citep{arazo2020pseudo, chen2022debiased, wang2022debiased} during training, ultimately resulting in suboptimal performance. 

\begin{figure*}
    \centering
    \begin{subfigure}{0.28\textwidth}
        \centering
        \includegraphics[width=\textwidth]{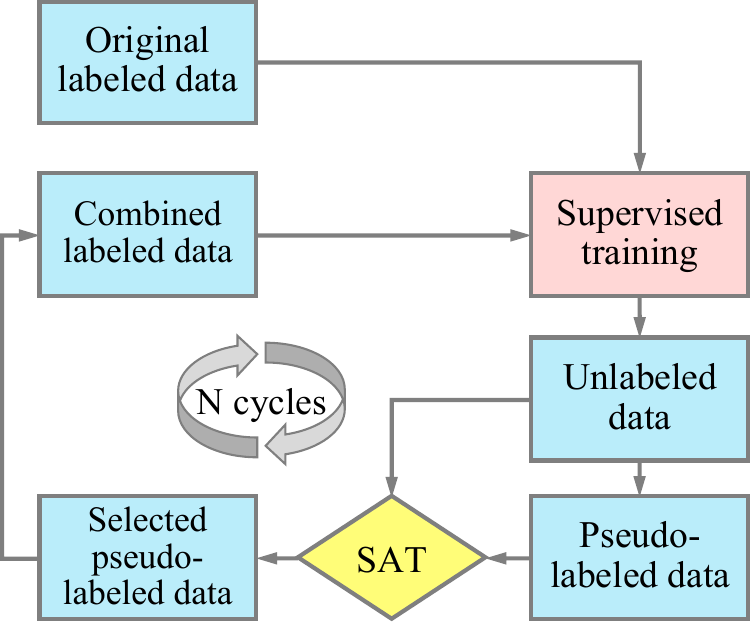}
        \caption{Super-SST}
        \label{fig:illustration_super}
    \end{subfigure}\hspace{10pt}
    \begin{subfigure}{0.66\textwidth}
        \centering
        \includegraphics[width=\textwidth]{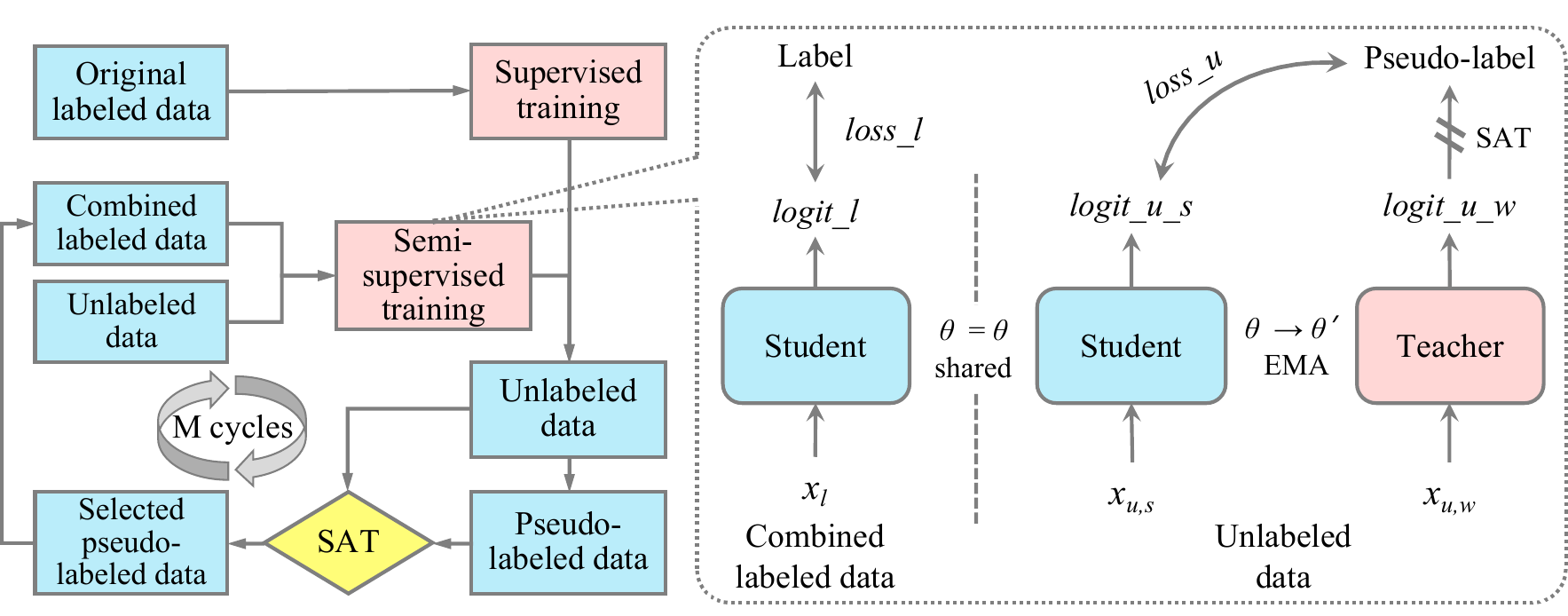}
        \caption{Semi-SST}
        \label{fig:illustration_semi}
    \end{subfigure}
    \caption{Illustration of the \textbf{S}elf-training with \textbf{S}elf-adaptive \textbf{T}hresholding (SST) framework. Our SST can be integrated with either supervised or semi-supervised algorithms, referred to as \textbf{Super-SST} and \textbf{Semi-SST}, respectively. Self-Adaptive Thresholding (SAT) is a core mechanism in our SST frameworks, accurately and adaptively adjusting class-specific thresholds according to the learning progress of the model. This ensures both high quality and sufficient quantity of selected pseudo-labeled data.}
    \label{fig:illustration}
\end{figure*}

To address the aforementioned issues, we propose Self-training with Self-adaptive Thresholding (SST), a novel, effective, and efficient SSL framework that significantly advances the field of SSL. As illustrated in Figure \ref{fig:illustration}, our SST framework can be integrated with either supervised or semi-supervised algorithms. We refer to the combination of supervised training and SST as \textit{Super-SST}, and the combination of semi-supervised training and SST as \textit{Semi-SST}. Our SST introduces an innovative Self-Adaptive Thresholding (SAT) mechanism, which accurately and adaptively adjusts class-specific thresholds according to the learning progress of the model. This approach ensures both high quality and sufficient quantity of selected pseudo-labeled data, mitigating the risks of inaccurate pseudo-labels and confirmation bias, thereby enhancing overall model performance. 

SAT differs from existing dynamic thresholding methods like FlexMatch and FreeMatch in three key aspects:
\begin{itemize}
    \item \textit{Threshold Derivation}: SAT adopts a confidence-based approach, applying a cutoff $C$ to filter out low-confidence probabilities per class, then averaging the remaining probabilities and scaling by a factor $S$ to derive class-specific thresholds. Unlike FlexMatch, which relies on curriculum learning based on predicted unlabeled sample counts and a global threshold, SAT avoids such dependencies. Unlike FreeMatch, which uses exponential moving averages (EMA) and a global threshold without low-confidence filtering, SAT introduces a cutoff mechanism while dispensing with both EMA and global threshold.
    
    \item \textit{Threshold Updates}: SAT updates thresholds once per training cycle using a well-trained model, limiting updates to single digits. This substantially reduces computational overhead and mitigates confirmation bias. In contrast, FlexMatch and FreeMatch update thresholds every iteration (e.g., over 1,000,000 updates) using a model still in training, incurring substantial computational cost and exacerbating confirmation bias during early training.

    \item \textit{Empirical Advantages}: Extensive empirical evaluations demonstrate SAT’s advantages across multiple benchmarks. For instance, on CIFAR-100 with 400, 2500, and 10000 labels, FreeMatch attains a mean error rate of 28.71\% using about 240 GPU hours. In contrast, Super-SST and Semi-SST, powered by SAT’s mechanism, achieve lower mean error rates of 22.74\% and 20.50\%, respectively, with only 3 and 4 GPU hours, yielding speedups of 80$\times$ and 60$\times$.
\end{itemize}

The Super-SST pipeline, illustrated in Figure \ref{fig:illustration_super}, consists of several steps: 1) Initialize the model with the original labeled data. 2) Utilize the well-trained, high-performance model to predict the unlabeled data. 3) Apply Self-Adaptive Thresholding (SAT) to accurately generate or adjust class-specific thresholds. 4) Select high-confidence and reliable pseudo-labels using the class-specific thresholds, reducing inaccurate pseudo-labels and mitigating confirmation bias. 5) Expand the labeled training set by combining the original labeled data with the carefully selected pseudo-labeled data, treating them equally. 6) Optimize the model’s performance using the combined labeled data. 7) Iterative training by returning to step 2, continuously refining the model over multiple cycles until its performance converges. 

As illustrated in Figure \ref{fig:illustration_semi}, Semi-SST enhances Super-SST’s offline pseudo-labeling pipeline by integrating an EMA-Teacher model \citep{cai2022semi}, introducing dynamic online pseudo-labeling to complement the offline process. Specifically, in EMA-Teacher, the teacher model dynamically generates pseudo-labels online, while the student model is trained on both offline combined labeled data and these online pseudo-labeled data. Semi-SST thus employs both offline and online pseudo-labeling mechanisms, with SAT applied in both cases to ensure the selection of high-confidence pseudo-labels. Table \ref{tab:super_semi_sst} outlines the primary distinctions between Super-SST and Semi-SST.

\begin{table}
    \caption{Comparison of Super-SST and Semi-SST.}
    \label{tab:super_semi_sst}
    \setlength{\tabcolsep}{8pt}
    \centering
    \footnotesize
    \renewcommand{\arraystretch}{1.5}
    \begin{tabular*}{\linewidth}{m{3cm}m{5.8cm}m{5.8cm}}
    \toprule
    Aspect                     & Super-SST                                                              & Semi-SST \\\midrule
    Core Training Algorithm    & Supervised training algorithm                                          & Semi-supervised training algorithm (EMA-Teacher) \\\hline
    Pseudo-labeling            & Offline only                                                           & Offline and online \\\hline
    SAT Application            & Offline pseudo-label selection                                         & Offline and online pseudo-label selection \\\hline
    Training Data              & Offline combined labeled data (original + SAT-selected pseudo-labels, treated as human labels)  & Offline combined labeled data and online pseudo-labeled data \\\hline
    Loss Function              & Cross-Entropy (CE) loss on offline combined labeled data ($\mathcal{L}_l$) & Total loss = CE loss on offline combined labeled data + weighted CE loss on online pseudo-labeled data ($\mathcal{L} = \mathcal{L}_l + \mu\mathcal{L}_u$) \\
    \bottomrule
    \end{tabular*}
\end{table}

Extensive experiments and results strongly confirm the effectiveness, efficiency, generalization, and scalability of our methods across both CNN-based and Transformer-based architectures and various datasets, including ImageNet-1K, CIFAR-100, Food-101, and iNaturalist. Compared to supervised and semi-supervised baselines, our methods achieve a relative improvement of 0.7\%-20.7\% (0.2\%-12.1\%) in Top-1 accuracy with only 1\% (10\%) labeled data. Notably, Semi-SST-ViT-Huge achieves the best results on competitive ImageNet-1K SSL benchmarks (no external data), with 80.7\% / 84.9\% Top-1 accuracy using only 1\% / 10\% labeled data. Compared to the fully-supervised DeiT-III-ViT-Huge, which achieves 84.8\% Top-1 accuracy using 100\% labeled data, our method demonstrates superior performance using only 10\% labeled data. This indicates a tenfold reduction in human annotation costs, significantly reducing the performance disparity between semi-supervised and fully-supervised methods. Moreover, our SST methods not only achieve high performance but also demonstrate remarkable efficiency, resulting in an exceptional cost-performance ratio. Furthermore, on the ImageNet-1K benchmark, Super-SST's accuracy outperforms FlexMatch and FreeMatch by 6.27\% and 4.99\%, respectively, requiring only 3 threshold updates, while FlexMatch and FreeMatch demand over 1,000,000 threshold updates.

In summary, our contributions are three-fold:
\begin{itemize}
    \item We propose an innovative Self-training with Self-adaptive Thresholding (SST) framework, along with its two variations: Super-SST and Semi-SST, significantly advancing the SSL field. 
    \item SST introduces a novel Self-Adaptive Thresholding (SAT) technique, which accurately and adaptively adjusts class-specific thresholds according to the model's learning progress. This mechanism ensures both the high quality and sufficient quantity of selected pseudo-labeled data, mitigating the risks of inaccurate pseudo-labels and confirmation bias, thereby enhancing overall model performance. 
    \item SST achieves state-of-the-art (SOTA) performance across various architectures and datasets while demonstrating remarkable efficiency, resulting in an exceptional cost-performance ratio. Moreover, SST significantly reduces the reliance on human annotation, bridging the gap between semi-supervised and fully-supervised learning. 
\end{itemize}

The organization of this paper is as follows: Section \ref{sec:research objectives} outlines the study's objectives. Section \ref{sec:methodology} details the methodology. Section \ref{sec:experiments and results} describes the experiments and reports the results. Section \ref{sec:related work} reviews related works, highlighting the differences between our methods and previous approaches. Section \ref{sec:limitation} and \ref{sec:conclusion} presents the limitation and conclusion, respectively.

\section{Research objectives}
\label{sec:research objectives}

The SSL methods previously discussed in Section \ref{sec:introduction} either rely on fixed thresholding or use costly and performance-limited dynamic thresholding techniques to select pseudo-labels. Both approaches tend to introduce more noisy labels during the early training stages, leading to a feedback loop of increasing inaccuracies and reinforcing confirmation bias, ultimately resulting in suboptimal performance. To overcome these challenges, we propose the SST framework and its core component, the SAT mechanism. These innovations aim to accurately select high-quality and sufficient pseudo-labeled data during training, minimizing the risks of inaccuracies and confirmation bias, thereby enhancing overall model performance. We hope this paper will inspire future research and provide cost-effective solutions for practical applications where acquiring labeled data is challenging or costly.

\section{Methodology}
\label{sec:methodology}

\begin{algorithm}[t]
\caption{PyTorch-like Pseudo-code of SST}
\label{alg:sst}
\definecolor{codeblue}{rgb}{0.25,0.5,0.5}
\lstset{
  backgroundcolor=\color{white},
  basicstyle=\fontsize{7.2pt}{7.2pt}\ttfamily\selectfont,
  columns=fullflexible,
  breaklines=true,
  captionpos=b,
  commentstyle=\fontsize{7.2pt}{7.2pt}\color{codeblue},
  keywordstyle=\fontsize{7.2pt}{7.2pt}
}
\begin{lstlisting}[language=python]
# Step 1: Initialize the model with original labeled data
model = initialize_model(labeled_data)

# Set hyperparameters
max_cycles = 6

for cycle in range(max_cycles):
    # Step 2: Predict unlabeled data with the well-trained high-performance model
    outputs = model.predict(unlabeled_data)
    probabilities = softmax(outputs)

    # Step 3: Generate class-specific thresholds with Self-Adaptive Thresholding (SAT)
    class_specific_thresholds = self_adaptive_thresholding(probabilities)

    # Step 4: Select reliable pseudo-labels with class-specific thresholds
    pseudo_labeled_data = []
    for data_point, probability in zip(unlabeled_data, probabilities):
        confidence, pseudo_label = max(probability)
        if confidence > class_specific_thresholds[pseudo_label]:
            pseudo_labeled_data.append((data_point, pseudo_label))
    
    # Step 5: Combine labeled data with selected pseudo-labeled data
    combined_labeled_data = labeled_data + pseudo_labeled_data

    # Step 6: Optimize the model's performance with the combined labeled data
    model.train(combined_labeled_data) # In Semi-SST, unlabeled data is also required here

    # Step 7: Return to step 2 for iterative training until the model's performance converges
    if model.has_converged():
        break

# Self-trained model is ready for inference
self_trained_model = model
\end{lstlisting}
\end{algorithm}

Figure \ref{fig:illustration} provides a visual overview of SST, while Algorithm \ref{alg:sst} presents its pseudo-code implementation. The algorithm is organized into several distinct steps: 

\textbf{1. Initialization}. We initialize the model parameters $\theta$ using the original labeled samples $\left\{\left(x_{li}, y_{li}\right)\right\}_{i=1}^{N_l}$. This initialization phase employs the standard Cross-Entropy (CE) loss Equation \ref{eq:loss_l} to optimize the model's initial parameters, ensuring a strong starting checkpoint for subsequent training phases.

\begin{equation}
   \mathcal{L}_l = \frac{1}{N_l} \sum_{i=1}^{N_l} \text{CE}\left(f(x_{li}; \theta), y_{li}\right)
   \label{eq:loss_l}
\end{equation}

\textbf{2. Prediction}. We use the well-trained high-performance model $\theta$ to predict all unlabeled data $\left\{x_{ui}\right\}_{i=1}^{N_u}$, outputting the probabilities over $N_{c}$ classes. 

\begin{equation}
   p_{ui} = \text{softmax}(f(x_{ui}; \theta)) \quad \text{for} \quad i = 1, 2, \ldots, N_u
\end{equation}

\noindent This step produces an $N_u \times N_{c}$ probability matrix, where $N_u$ represents the number of unlabeled samples and $N_{c}$ denotes the total number of classes. Each element in this matrix corresponds to the predicted probability of a sample belonging to a particular class.

\textbf{3. Self-Adaptive Thresholding (SAT)}. With the generated $N_u \times N_{c}$ probability matrix, we proceed to apply the SAT mechanism. The SAT mechanism is a pivotal component of the SST framework, designed to accurately and adaptively adjust class-specific thresholds according to the learning progress of the model. SAT mechanism involves several sub-steps. 

1) Sorting: Given the $N_u \times N_{c}$ probability matrix \(\mathbf{P}\), where \(\mathbf{P}_{i,j}\) indicates the predicted probability of the \(i\)-th unlabeled sample belonging to the \(j\)-th class, we sort the probabilities for each class \(j\) in descending order. Let \(\mathbf{P}_{:,j}\) denote the vector of probabilities for class \(j\).

\begin{equation}
    \mathbf{P}_{:,j}^{\text{sorted}} = \text{sort}(\mathbf{P}_{:,j}, \text{descending})
\end{equation}

2) Cutoff Filtering: For each class \(j\), we filter out probabilities that are lower than a predefined cutoff value \(C\) (\eg, 0.5), ensuring that only higher, reliable probabilities are retained. Let \(\mathbf{P}_{:,j}^{\text{filtered}}\) denote the filtered probabilities for class \(j\):

\begin{equation}
    \mathbf{P}_{:,j}^{\text{filtered}} = \{p \in \mathbf{P}_{:,j}^{\text{sorted}} \mid p > C\}
\end{equation}

3) Average Calculation: We compute the average of the remaining probabilities for each class \(j\): 

\begin{equation}
    \bar{p}_j = \frac{1}{|\mathbf{P}_{:,j}^{\text{filtered}}|} \sum_{p \in \mathbf{P}_{:,j}^{\text{filtered}}} p
\end{equation}

\noindent where \(|\mathbf{P}_{:,j}^{\text{filtered}}|\) is the number of filtered probabilities for class \(j\).

4) Threshold Scaling: We then multiply the average value of each class \(j\) by a scaling factor \(S\) to obtain the class-specific thresholds \(\tau_j\):

\begin{equation}
    \tau_j = S \cdot \bar{p}_j \quad \text{for} \quad j = 1, 2, \ldots, N_{c}
\end{equation}

The SAT mechanism ensures that lower, less reliable probabilities do not influence the threshold computation, thereby maintaining the reliability and quality of the pseudo-labeling process.

\textbf{4. Selection}. We use the class-specific thresholds $\left\{\tau_{j}\right\}_{j=1}^{N_{c}}$ to select high-confidence and reliable pseudo-labels, reducing the introduction of inaccurate pseudo-labels and mitigating confirmation bias. For each unlabeled sample $x_u \in \left\{x_{ui}\right\}_{i=1}^{N_u}$, the pseudo-label is produced by $\hat{y}_{ui} = \arg \max_j p_{ui,j}$ with its associated confidence $\hat{p}_{ui} = \max_j p_{ui,j}$. An unlabeled sample $x_{ui}$ is selected as a pseudo-labeled data point if $\hat{p}_{ui} > \tau_{\hat{y}_{ui}}$. 

\begin{equation}
   \hat{y}_{ui} = \arg\max_j p_{ui,j}
\end{equation}

\begin{equation}
   \hat{p}_{ui} = \max_j p_{ui,j}
\end{equation}

\textbf{5. Combination}. We combine the labeled data with the carefully selected high-confidence pseudo-labeled data, thereby expanding the labeled training set. 

\begin{equation}
   \{(x_i, y_i)\}_{i=1}^{N_{l+u}} = \{(x_{li}, y_{li})\}_{i=1}^{N_l} \cup \{(x_{ui}, \hat{y}_{ui}) \mid \hat{p}_{ui} > \tau_{\hat{y}_{ui}}\}_{i=1}^{N_u}
\end{equation}

\textbf{6. Optimization}. In the Super-SST framework, the model is optimized on the offline combined labeled data using the standard CE loss, as defined in Equation \ref{eq:loss_l}. As discussed in Section \ref{sec:introduction}, the high-confidence pseudo-labeled data included in the combined labeled data are selected offline via SAT and are treated as equally reliable as human-labeled data. In contrast, Semi-SST is built upon an EMA-Teacher framework, where the teacher model dynamically generates pseudo-labels during training, and the student model is trained on both offline combined labeled data and online pseudo-labeled data. We optimize the student model’s performance using the total loss function \(\mathcal{L}\), which consists of the CE loss for offline combined labeled data \(\mathcal{L}_l\) and the CE loss for online pseudo-labeled data \(\mathcal{L}_u\), weighted by a factor \(\mu\) to control the contribution of the \(\mathcal{L}_u\). This formulation ensures that the model benefits from both high-confidence offline pseudo-labels and dynamic online pseudo-labeling, with SAT filtering applied in both cases to maintain pseudo-label reliability. 

\begin{equation}
    \mathcal{L}=\mathcal{L}_l + \mu \mathcal{L}_u,
    \label{eq:loss_combined}
\end{equation}

\begin{equation}
   \mathcal{L}_u = \frac{1}{N_u} \sum_{i=1}^{N_u} \left[\hat{p}_{ui} > \tau_{\hat{y}_{ui}}\right] \cdot \text{CE}\left(f(x_{ui}; \theta), \hat{y}_{ui}\right),
   \label{eq:loss_u}
\end{equation}

\noindent where [$\cdot$] is the indicator function, and $\tau_{\hat{y}_{ui}}$ is the threshold of class $\hat{y}_{ui}$. 

As illustrated in Figure \ref{fig:illustration_semi}, for an unlabeled sample $x_u$, both weak and strong augmentations are applied, generating $x_{u,w}$ and $x_{u,s}$, respectively. The weakly-augmented $x_{u,w}$ is processed by the teacher network, outputting probabilities over classes. The pseudo-label is generated as described previously. The pseudo-labels selected by SAT can be used to guide the student model's learning on the strongly-augmented $x_{u,s}$, minimizing the CE loss. In Semi-SST's semi-supervised training, we utilize the EMA-Teacher algorithm, updating the teacher model parameters $\theta^{\prime}$ through exponential moving average (EMA) from the student model parameters $\theta$: 

\begin{equation}
    \theta^{\prime} := m \theta^{\prime} + (1 - m) \theta,
    \label{eq:ema}
\end{equation}

\noindent where the momentum decay $m$ is a number close to 1, such as 0.9999. 

\textbf{7. Iterative training}. Return to the prediction step for iterative training, continuously refining the model over multiple cycles until its performance converges. More details of the pipeline can be found in Figure \ref{fig:illustration} and Algorithm \ref{alg:sst}.

\section{Experiments and results}
\label{sec:experiments and results}

\subsection{Datasets and implementation details}

\textbf{Datasets}. As shown in Table \ref{tab:datasets}, we conduct experiments on eight publicly accessible benchmark datasets: ImageNet-1K, Food-101, iNaturalist, CIFAR-100, CIFAR-10, SVHN, STL-10, and Clothing-1M. These datasets collectively provide a comprehensive benchmark for evaluating the performance of SSL algorithms across a wide variety of domains and complexities. All SSL experiments in this paper are performed on randomly sampled labeled data from the training set of each dataset.
\begin{itemize}
    \item ImageNet-1K \citep{russakovsky2015imagenet} covers a broad spectrum of real-world objects, which is extensively used in computer vision tasks. It contains approximately 1.28M training images and 50K validation images, divided into 1,000 categories.
    \item Food-101 \citep{bossard2014food} consists of 101K food images across 101 categories. Each class contains 1,000 images, divided into 750 training images and 250 test images.
    \item iNaturalist \citep{inaturalist-2019-fgvc6} is derived from the iNaturalist community, which documents biodiversity across the globe. It contains 256,213 training images and 3,030 test images, covering 1,010 classes of living organisms.
    \item CIFAR-100 \citep{krizhevsky2009learning} covers a diverse range of real-world objects and animals. It contains 60K 32$\times$32 images divided into 100 categories, with each class containing 500 training images and 100 testing images.
    \item CIFAR-10 \citep{krizhevsky2009learning} comprises 60K 32$\times$32 color images divided into 10 distinct categories, with each class containing 5,000 training images and 1,000 testing images.
    \item SVHN \citep{netzer2011reading} is a real-world dataset for digit recognition obtained from house numbers in Google Street View images. It contains 73,257 training samples and 26,032 testing samples across 10 digit classes, with an additional 531,131 supplementary images.
    \item STL-10 \citep{coates2011analysis} consists of 13K 96$\times$96 color images divided into 10 classes. Each category includes 500 labeled training images and 800 testing images, supplemented by 100K unlabeled images.
    \item Clothing-1M \citep{xiao2015learning} is a large-scale dataset containing approximately 1 million images of clothing collected from online shopping websites. The images are annotated with noisy labels across 14 categories, and the dataset is split into 47,570 clean labeled training images, 1M noisy labeled training images, 14,313 validation images and 10,526 testing images, making it a challenging benchmark for learning with noisy labels.
\end{itemize}

\textbf{Architectures}. For transformer neural networks, we utilize the standard Vision Transformer (ViT) architectures \citep{dosovitskiy2020image}, which consist of a stack of Transformer blocks. Each block includes a multi-head self-attention layer and a multi-layer perceptron (MLP) block. We use average pooling of the ViT output for classification. Following Semi-ViT, we directly use the pretrained ViT-Small from DINO \citep{caron2021emerging}, ViT-Base and ViT-Huge from MAE \citep{he2022masked}, which have learned useful visual representations for downstream tasks. For convolutional neural networks, we utilize the standard ConvNeXt architecture \citep{liu2022convnet}, which incorporates mixup strategy \citep{zhang2017mixup} for enhanced results, instead of the traditional ResNet \citep{he2016deep}. Note that self-supervised pretraining is optional in our methods. We train the ConvNeXt architecture from scratch without any pretraining.

\textbf{Implementation details}. Our training protocol largely follows the good practice in Semi-ViT, with minor differences. The optimization of model is performed using AdamW \citep{loshchilov2017decoupled}, with linear learning rate scaling rule \citep{goyal2017accurate} $lr = base\_lr \times batchsize/256$ and cosine learning rate decay schedule \citep{loshchilov2016sgdr}. Additionally, we incorporate label smoothing \citep{szegedy2016rethinking}, drop path \citep{huang2016deep}, and cutmix \citep{yun2019cutmix} in our experiments. For fair comparisons, we use a center crop of 224$\times$224 during training and testing. The data augmentation strategy follows the methods described in \citep{cai2022semi, cubuk2020randaugment, zhong2020random}. Detailed hyper-parameter settings are provided in Table \ref{tab:settings}.

\begin{table}
    \caption{All experiments in this paper are performed on eight publicly accessible benchmark datasets.}
    \label{tab:datasets}
    \setlength{\tabcolsep}{32pt}
    \centering
    \small
    \begin{tabular*}{\linewidth}{lrrr}
        \toprule
        Dataset& \#Training data& \#Test data& \#Class\\\midrule
        ImageNet-1K \citep{russakovsky2015imagenet} & 1,281,167& 50,000& 1,000\\
        iNaturalist \citep{inaturalist-2019-fgvc6} &    265,213&  3,030& 1,010\\
        Food-101 \citep{bossard2014food} &               75,750& 25,250&   101\\
        CIFAR-100 \citep{krizhevsky2009learning} &       50,000& 10,000&   100\\
        CIFAR-10 \citep{krizhevsky2009learning} &        50,000& 10,000&   10\\
        SVHN \citep{netzer2011reading} &                 73,257& 26,032&   10\\
        STL-10 \citep{coates2011analysis} &               5,000&  8,000&   10\\
        Clothing-1M \citep{xiao2015learning} &        1,047,570& 10,526&   14\\
        \bottomrule
    \end{tabular*}
\end{table}

\subsection{Comparison with baselines}
\label{sec:comparison with baselines}

\begin{table}
    \caption{Comparison with both supervised and semi-supervised baselines across various datasets and architectures. The performance gaps to the baselines are indicated in brackets. The results of supervised baseline are implemented by us. Semi-ViT did not conduct experiments on CIFAR-100, resulting in the absence of corresponding results.}
    \label{tab:comparison with baselines}
    \setlength{\tabcolsep}{13pt}
    \centering
    \begin{tabular*}{\linewidth}{llllll}
        \toprule
        Dataset& Architecture& Parameter& Method& 1\%& 10\%\\\hline
        \multirow{12}{*}{ImageNet-1K}& \multirow{4}{*}{ViT-Small}& \multirow{4}{*}{22M}& Supervised& 60.3& 74.3\\
                                     &                           &                     & Super-SST (ours)& 70.4 (\textcolor{hscolor}{+10.1})& 78.3 (\textcolor{hscolor}{+4.0})\\\cline{4-6}
                                     &                           &                     & Semi-ViT& 68.0& 77.1\\
                                     &                           &                     & Semi-SST (ours)& 71.4 (\textcolor{hscolor}{+3.4})& 78.6 (\textcolor{hscolor}{+1.5})\\\cline{2-6}
                                     & \multirow{4}{*}{ViT-Huge}& \multirow{4}{*}{632M}& Supervised& 73.7& 81.8\\
                                     &                           &                     & Super-SST (ours)& 80.3 (\textcolor{hscolor}{+6.6})& 84.8 (\textcolor{hscolor}{+3.0})\\\cline{4-6}
                                     &                           &                     & Semi-ViT& 80.0& 84.3\\
                                     &                           &                     & Semi-SST (ours)& 80.7 (\textcolor{hscolor}{+0.7})& 84.9 (\textcolor{hscolor}{+0.6})\\\cline{2-6}
                                     & \multirow{4}{*}{ConvNeXt-T}& \multirow{4}{*}{28M}& Supervised& -& 65.8\\
                                     &                           &                     & Super-SST (ours)& -& 77.9 (\textcolor{hscolor}{+12.1})\\\cline{4-6}
                                     &                           &                     & Semi-ViT& -& 74.1\\
                                     &                           &                     & Semi-SST (ours)& -& 78.6 (\textcolor{hscolor}{+4.5})\\\hline
        \multirow{4}{*}{CIFAR-100}   & \multirow{4}{*}{ViT-Small}& \multirow{4}{*}{22M}& Supervised& 53.7& 79.8\\
                                     &                           &                     & Super-SST (ours)& 68.1 (\textcolor{hscolor}{+14.4})& 84.1 (\textcolor{hscolor}{+4.3})\\\cline{4-6}
                                     &                           &                     & Semi-ViT& -& -\\
                                     &                           &                     & Semi-SST (ours)& 72.3& 84.7\\\hline
        \multirow{4}{*}{Food-101}    & \multirow{4}{*}{ViT-Base}& \multirow{4}{*}{86M}& Supervised& 62.7& 84.6\\
                                     &                           &                     & Super-SST (ours)& 83.4 (\textcolor{hscolor}{+20.7})& 91.1 (\textcolor{hscolor}{+6.5})\\\cline{4-6}
                                     &                           &                     & Semi-ViT& 82.1& 91.3\\
                                     &                           &                     & Semi-SST (ours)& 86.5 (\textcolor{hscolor}{+4.4})& 91.5 (\textcolor{hscolor}{+0.2})\\\hline
        \multirow{4}{*}{iNaturalist} & \multirow{4}{*}{ViT-Base}& \multirow{4}{*}{86M}& Supervised& 22.6& 57.5\\
                                     &                           &                     & Super-SST (ours)& 25.6 (\textcolor{hscolor}{+3.0})& 65.3 (\textcolor{hscolor}{+7.8})\\\cline{4-6}
                                     &                           &                     & Semi-ViT& 32.3& 67.7\\
                                     &                           &                     & Semi-SST (ours)& 33.4 (\textcolor{hscolor}{+1.1})& 68.1 (\textcolor{hscolor}{+0.4})\\
        \bottomrule
    \end{tabular*}
\end{table}

Table \ref{tab:comparison with baselines} presents a comparison of Super-SST against supervised baselines and Semi-SST against semi-supervised baselines. This comparison covers both CNN-based and Transformer-based architectures across various datasets, including ImageNet-1K, CIFAR-100, Food-101, and iNaturalist, with 1\% and 10\% labeled data. Due to computational resource constraints, we only performed experiments with ConvNeXt-T on 10\% ImageNet-1K labels.

\textbf{Effectiveness and generalization}. As shown in Table \ref{tab:comparison with baselines}, both Super-SST and Semi-SST consistently outperform their respective supervised and semi-supervised baselines across various datasets, demonstrating strong generalization capabilities. Notably, both CNN and Transformer architectures benefit significantly from Super-SST and Semi-SST, underscoring their effectiveness and robustness across different architectures.

\textbf{Data Efficiency}. Table \ref{tab:comparison with baselines} also demonstrates that both Super-SST and Semi-SST maintain robust performance improvements across different percentages of labeled data. They achieve a relative improvement of 0.7\%-20.7\% with 1\% labeled data and 0.2\%-12.1\% with 10\% labeled data in Top-1 accuracy. The improvements are more pronounced with a smaller percentage of labeled data (\eg, 1\%), indicating that Super-SST and Semi-SST are particularly effective in low-data regimes, making them suitable for data-efficient learning. As the amount of labeled data increases from 1\% to 10\%, the relative improvements decrease, indicating diminishing returns with more labeled data. 

\textbf{Scalability}. As presented in Table \ref{tab:comparison with baselines}, both Super-SST and Semi-SST show significant improvements with larger architectures like ViT-Huge, indicating better scalability with increased model capacity. For example, with 1\% / 10\% ImageNet-1K labeled data and ViT-Huge, Super-SST shows a 6.6\% / 3.0\% relative improvement over the supervised baseline, while Semi-SST shows a 0.7\% / 0.6\% relative improvement over the semi-supervised baseline. 

\textbf{Super-SST vs. Semi-SST}. Both Super-SST and Semi-SST demonstrate competitive performance across various datasets and architectures, with Semi-SST generally having a slight edge. As shown in Table \ref{tab:comparison with baselines}, Semi-SST consistently achieves higher Top-1 accuracy compared to Super-SST. However, Super-SST typically delivers a larger relative improvement over its baselines. For example, using 1\% ImageNet-1K labeled data and ViT-Small, Super-SST / Semi-SST achieves 70.4\% / 71.4\% Top-1 accuracy, while the relative improvement over supervised / semi-supervised baseline is 10.1\% / 3.4\%. 

In summary, Table \ref{tab:comparison with baselines} demonstrates that Super-SST and Semi-SST offer significant performance improvements across various datasets and architectures, particularly excelling in low-data scenarios and scaling well with model capacity. While Semi-SST often achieves higher Top-1 accuracy compared to Super-SST, Super-SST generally delivers larger improvement over its baselines. Our methods are effective, scalable, and broadly applicable, making them valuable for practical applications where labeled data is limited.

\subsection{Comparison with SOTA SSL models}
\label{sec:comparison with sota SSL models}

\begin{table}
    \caption{Comparison with previous SOTA SSL methods across both CNN-based and Transformer-based architectures on 1\% and 10\% ImageNet-1K labeled data. Our Super-SST and Semi-SST methods significantly outperform previous SOTA SSL counterparts.}
    \label{tab:comparison with sota SSL methods}
    \setlength{\tabcolsep}{20.4pt}
    \centering
    \small
    \begin{tabular*}{\linewidth}{lllrcc}
        \toprule
        \multicolumn{2}{l}{\multirow{2}{*}{\quad\quad Method}}& \multirow{2}{*}{Architecture}& \multirow{2}{*}{Param}& \multicolumn{2}{c}{ImageNet-1K}\\
        &                          &                      &    & 1\%& 10\%\\\hline
        \multirow{13}{*}{\rotatebox[origin=c]{90}{CNN}}
        &UDA \citep{xie2020unsupervised}&        ResNet-50& 26M& -& 68.8\\
        &SimCLRv2 \citep{chen2020big}&           ResNet-50& 26M& 60.0& 70.5\\
        &FixMatch \citep{sohn2020fixmatch}&      ResNet-50& 26M& -& 71.5\\
        &MPL \citep{pham2021meta}&               ResNet-50& 26M& -& 73.9\\
        &EMAN \citep{cai2021exponential}&        ResNet-50& 26M& 63.0& 74.0\\
        &CoMatch \citep{li2021comatch}&          ResNet-50& 26M& 66.0& 73.6\\
        &PAWS \citep{assran2021semi}&            ResNet-50& 26M& 66.5& 75.5\\
        &SimMatch \citep{zheng2022simmatch}&     ResNet-50& 26M& 67.2& 74.4\\
        &SimMatchV2 \citep{zheng2023simmatchv2}& ResNet-50& 26M& 71.9& 76.2\\
        &SimMatchV2 \citep{zheng2023simmatchv2}& ConvNeXt-T& 28M& 71.3& 77.8\\
        &Semi-ViT \citep{cai2022semi}&           ConvNeXt-T& 28M& -& 74.1\\\cline{2-6}
        &Super-SST (ours)&                       ConvNeXt-T& 28M& -& 77.9\\
        &Semi-SST (ours)&                        ConvNeXt-T& 28M& -& 78.6\\\hline
        \multirow{10}{*}{\rotatebox[origin=c]{90}{Transformer}}
        &SemiFormer \citep{weng2022semi}        & ViT-S+Conv& 42M&  -& 75.5\\
        &SimMatchV2 \citep{zheng2023simmatchv2}& ViT-Small& 22M& 63.7& 75.9\\
        &Semi-ViT \citep{cai2022semi}&           ViT-Small& 22M& 68.0& 77.1\\
        &Semi-ViT \citep{cai2022semi}&           ViT-Base&  86M& 71.0& 79.7\\
        &Semi-ViT \citep{cai2022semi}&           ViT-Large& 307M& 77.3& 83.3\\
        &Semi-ViT \citep{cai2022semi}&           ViT-Huge& 632M& 80.0& 84.3\\\cline{2-6}
        &Super-SST (ours)&                       ViT-Small& 22M& 70.4& 78.3\\
        &Super-SST (ours, distilled)&            ViT-Small& 22M& 76.9& 80.3\\
        &Semi-SST (ours)&                        ViT-Small& 22M& 71.4& 78.6\\
        &Super-SST (ours)&                       ViT-Huge& 632M& 80.3& 84.8\\
        &Semi-SST (ours)&                        ViT-Huge& 632M& \textbf{80.7}& \textbf{84.9}\\
        \bottomrule
    \end{tabular*}
\end{table}

Table \ref{tab:comparison with sota SSL methods} presents a comprehensive comparison between our methods and previous SOTA SSL methods, including UDA \citep{xie2020unsupervised}, SimCLRv2 \citep{chen2020big}, FixMatch \citep{sohn2020fixmatch}, MPL \citep{pham2021meta}, EMAN \citep{cai2021exponential}, CoMatch \citep{li2021comatch}, PAWS \citep{assran2021semi}, SimMatch \citep{zheng2022simmatch}, SimMatchV2 \citep{zheng2023simmatchv2}, and Semi-ViT \citep{cai2022semi}, focusing on their performance with 1\% and 10\% ImageNet-1K labeled data. Our methods are evaluated across both CNN-based and Transformer-based architectures. Following Semi-ViT, we utilize the recently proposed ConvNeXt instead of the traditional ResNet.

\textbf{Superiority and robustness}. For CNN-based models, Super-SST and Semi-SST achieve 77.9\% and 78.6\% Top-1 accuracy, respectively, with ConvNeXt-T on 10\% ImageNet-1K labeled data, surpassing all previous counterparts. For Transformer-based models, Super-SST / Semi-SST achieves 70.4\% / 71.4\% Top-1 accuracy with ViT-Small on 1\% labeled data, and 78.3\% / 78.6\% on 10\% labeled data, significantly outperforming previous counterparts. Both Super-SST and Semi-SST consistently outperform their counterparts across CNN-based and Transformer-based architectures, highlighting the superiority and robustness of our methods. 

\textbf{Scalability and data-efficiency}. Scalable and data-efficient algorithms are crucial to semi-supervised learning. As demonstrated in Table \ref{tab:comparison with sota SSL methods}, both Super-SST and Semi-SST can easily scale up to larger architectures like ViT-Huge. Notably, Semi-SST-ViT-Huge achieves the best 80.7\% / 84.9\% Top-1 accuracy on 1\% / 10\% labeled data, setting new SOTA performance on the highly competitive ImageNet-1K SSL benchmarks without external data. Moreover, larger architectures like ViT-Huge generally achieve higher accuracy compared to smaller architectures, demonstrating that larger architectures are strong data-efficient learners. This trend is consistent in both 1\% and 10\% labeled data scenarios, underscoring the potential for larger models to better utilize scarce labeled data. 

\textbf{Knowledge distillation}. We have shown that larger architectures like ViT-Huge generally achieve higher accuracy than smaller ones, particularly in low-data scenarios. However, large models with high capacity require significant computational resources during deployment, making them impractical and uneconomical for real-world applications. Conversely, small models are easier to deploy but tend to perform worse compared to large models. This raises an important question: how can we train a small, high-performance model suitable for practical deployment in low-data scenarios? To address this, we conduct huge-to-small knowledge distillation \citep{hinton2015distilling, mirzadeh2020improved, chen2021cross, passalis2018learning, park2019relational, zhu2021complementary, yang2022mixskd} experiments. Initially, we train a teacher model (\ie, ViT-Huge) using Super-SST on 1\% / 10\% original labeled data, achieving 80.3\% / 84.8\% Top-1 accuracy. We then use the well-trained teacher model and SAT technique to obtain selected pseudo-labeled data. Finally, we distill the knowledge from the teacher model into the student model (\ie, ViT-Small). The student model was supervised trained on a combination of few original labeled data and selected pseudo-labeled data, achieving 76.9\% and 80.3\% Top-1 accuracy in 1\% and 10\% low-data scenarios, respectively. This represents a substantial 6.5\% and 2.0\% relative improvement over the Super-SST-ViT-Small. Thus, we successfully develop a compact, lightweight and high-performance model that is well-suited for practical deployment. 

\begin{table}[t!]
\centering
\caption{Error Rates (\%) and Rankings on Classical SSL Benchmarks. The experimental settings in this table are consistent with those of SimMatchV2 \citep{zheng2023simmatchv2} and USB \citep{wang2022usb}, with all non-ours results directly sourced from these studies. Error rates and standard deviations are reported as the average over three runs. For each experimental setting, the best result is highlighted in bold, while the second-best is underlined.}
\label{tab:classic-cv-results}
\resizebox{\textwidth}{!}{
\begin{tabular}{l|ccc|ccc|ccc|ccc|c|c|c}
\toprule
Dataset & \multicolumn{3}{c|}{CIFAR-10}& \multicolumn{3}{c|}{CIFAR-100} & \multicolumn{3}{c|}{SVHN} & \multicolumn{3}{c|}{STL-10} & \multicolumn{1}{c|}{Friedman}   & \multicolumn{1}{c|}{Final} & \multicolumn{1}{c}{Mean} \\ \cmidrule(r){1-1}\cmidrule(lr){2-4}\cmidrule(lr){5-7}\cmidrule(lr){8-10}\cmidrule(l){11-13}
\# Label & \multicolumn{1}{c}{40} & \multicolumn{1}{c}{250} & \multicolumn{1}{c|}{4000}  & \multicolumn{1}{c}{400} & \multicolumn{1}{c}{2500} & \multicolumn{1}{c|}{10000} & \multicolumn{1}{c}{40} & \multicolumn{1}{c}{250}  & \multicolumn{1}{c|}{1000} & \multicolumn{1}{c}{40} & \multicolumn{1}{c}{250}  & \multicolumn{1}{c|}{1000} & \multicolumn{1}{c|}{rank} & \multicolumn{1}{c|}{rank} & \multicolumn{1}{c}{error rate} \\ \cmidrule(r){1-1}\cmidrule(lr){2-4}\cmidrule(lr){5-7}\cmidrule(lr){8-10}\cmidrule(lr){11-13}\cmidrule(lr){14-14}\cmidrule(lr){15-15}\cmidrule(lr){16-16}
Fully-Supervised &   \multicolumn{3}{c|}{4.57\tiny{±0.06}}                    & \multicolumn{3}{c|}{18.96\tiny{±0.06}}    & \multicolumn{3}{c|}{2.14\tiny{±0.01}}         & \multicolumn{3}{c|}{-}     &       -          &        -              &  -  \\ 
Supervised      & 77.18\tiny{±1.32}         & 56.24\tiny{±3.41}         & 16.10\tiny{±0.32}         & 89.60\tiny{±0.43}          & 58.33\tiny{±1.41}          & 36.83\tiny{±0.21}          & 82.68\tiny{±1.91}         & 24.17\tiny{±1.65}         & 12.19\tiny{±0.23}         & 75.4\tiny{±0.66}           & 55.07\tiny{±1.83}         & 35.42\tiny{±0.48}         &   -   &  -  &  -  \\
\cmidrule(r){1-1}\cmidrule(lr){2-4}\cmidrule(lr){5-7}\cmidrule(lr){8-10}\cmidrule(lr){11-13}\cmidrule(lr){14-14}\cmidrule(lr){15-15}\cmidrule(lr){16-16}
$\Pi$-Model     & 76.35\tiny{±1.69}         & 48.73\tiny{±1.07}         & 13.63\tiny{±0.07}         & 87.67\tiny{±0.79}          & 56.40\tiny{±0.69}          & 36.73\tiny{±0.05}          & 80.07\tiny{±1.22}         & 13.46\tiny{±0.61}         & 6.90\tiny{±0.22}          & 74.89\tiny{±0.57}          & 52.20\tiny{±2.11}         & 31.34\tiny{±0.64}         &16.17  &17  &48.20 \\
Pseudo-Labeling & 75.95\tiny{±1.86}         & 51.12\tiny{±2.91}         & 15.32\tiny{±0.35}         & 88.18\tiny{±0.89}          & 55.37\tiny{±0.48}          & 36.58\tiny{±0.12}          & 75.98\tiny{±5.36}         & 16.47\tiny{±0.59}         & 9.37\tiny{±0.42}          & 74.02\tiny{±0.47}          & 51.90\tiny{±1.87}         & 30.77\tiny{±0.04}         &16.08  &16  &48.42 \\
Mean Teacher    & 72.42\tiny{±2.10}         & 37.56\tiny{±4.90}         & 8.29\tiny{±0.10}          & 79.96\tiny{±0.53}          & 44.37\tiny{±0.60}          & 31.39\tiny{±0.11}          & 49.34\tiny{±7.90}         & 3.44\tiny{±0.02}          & 3.28\tiny{±0.07}          & 72.90\tiny{±0.83}          & 49.30\tiny{±2.09}         & 27.92\tiny{±1.65}         &13.67  &14  &40.01 \\
VAT             & 78.58\tiny{±2.78}         & 28.87\tiny{±3.62}         & 10.90\tiny{±0.16}         & 83.60\tiny{±4.21}          & 46.20\tiny{±0.80}          & 32.14\tiny{±0.31}          & 84.12\tiny{±3.18}         & 3.38\tiny{±0.12}          & 2.87\tiny{±0.18}          & 73.33\tiny{±0.47}          & 57.78\tiny{±1.47}         & 40.98\tiny{±0.96}         &14.67  &15  &45.23 \\
MixMatch        & 35.18\tiny{±3.87}         & 13.00\tiny{±0.80}         & 6.55±\tiny{0.05}          & 64.91\tiny{±3.34}          & 39.29\tiny{±0.13}          & 27.74\tiny{±0.27}          & 27.77\tiny{±5.43}         & 4.69\tiny{±0.46}          & 3.85\tiny{±0.28}          & 49.84\tiny{±0.58}          & 32.05\tiny{±1.16}         & 20.17\tiny{±0.67}         &12.92  &13  &27.09 \\
ReMixMatch      & 8.13\tiny{±0.58}          & 6.34\tiny{±0.22}          & 4.65\tiny{±0.09}          & 41.60\tiny{±1.48}          & 25.72\tiny{±0.07}          & 20.04\tiny{±0.13}          & 16.43\tiny{±13.77}        & 5.65\tiny{±0.35}          & 5.36\tiny{±0.58}          & 27.87\tiny{±3.85}          & 11.14\tiny{±0.52}         & 6.44\tiny{±0.15}          &9.50   &12  &14.95 \\
UDA             & 10.01\tiny{±3.34}         & 5.23\tiny{±0.08}          & 4.36\tiny{±0.09}          & 45.48\tiny{±0.37}          & 27.51\tiny{±0.28}          & 23.12\tiny{±0.45}          & 5.28\tiny{±4.02}          & \textbf{1.93}\tiny{±0.03}          & \textbf{1.94}\tiny{±0.02}          & 40.09\tiny{±4.03}          & 10.11\tiny{±1.15}         & 6.23\tiny{±0.28}          &8.00   &10  &15.11 \\
FixMatch        & 12.66\tiny{±4.49}         & 4.95\tiny{±0.10}          & 4.26\tiny{±0.01}          & 45.38\tiny{±2.07}          & 27.71\tiny{±0.42}          & 22.06\tiny{±0.10}          & \underline{3.37}\tiny{±1.01}          & \underline{1.97}\tiny{±0.01}          & 2.02\tiny{±0.03}          & 38.19\tiny{±4.76}          & 8.64\tiny{±0.84}          & 5.82\tiny{±0.06}          &6.83   &7   &14.75 \\
Dash            & 9.29\tiny{±3.28}          & 5.16\tiny{±0.28}          & 4.36\tiny{±0.10}          & 47.49\tiny{±1.05}          & 27.47\tiny{±0.38}          & 21.89\tiny{±0.16}          & 5.26\tiny{±2.02}          & 2.01\tiny{±0.01}          & 2.08\tiny{±0.09}          & 42.00\tiny{±4.94}          & 10.50\tiny{±1.37}         & 6.30\tiny{±0.49}          &8.25   &11  &15.32 \\
CoMatch         & 6.51\tiny{±1.18}          & 5.35\tiny{±0.14}          & 4.27\tiny{±0.12}          & 53.41\tiny{±2.36}          & 29.78\tiny{±0.11}          & 22.11\tiny{±0.22}          & 8.20\tiny{±5.32}          & 2.16\tiny{±0.04}          & 2.01\tiny{±0.04}          & 13.74\tiny{±4.20}          & 7.63\tiny{±0.94}          & 5.71\tiny{±0.08}          &7.17   &8   &13.41 \\
CRMatch         & 13.62\tiny{±2.62}         & 4.61\tiny{±0.17}          & 3.65\tiny{±0.04}          & 37.76\tiny{±1.45}          & 24.13\tiny{±0.16}          & 19.89\tiny{±0.23}          & \textbf{2.60}\tiny{±0.77}          & 1.98\tiny{±0.04}          & \underline{1.95}\tiny{±0.03}          & 33.73\tiny{±1.17}          & 14.87\tiny{±5.09}         & 6.53\tiny{±0.36}          &5.67   &4   &13.78 \\
FlexMatch       & 5.29\tiny{±0.29}          & 4.97\tiny{±0.07}          & 4.24\tiny{±0.06}          & 40.73\tiny{±1.44}          & 26.17\tiny{±0.18}          & 21.75\tiny{±0.15}          & 5.42\tiny{±2.83}          & 8.74\tiny{±3.32}          & 7.90\tiny{±0.30}          & 29.12\tiny{±5.04}          & 9.85\tiny{±1.35}          & 6.08\tiny{±0.34}          &7.58   &9   &14.19 \\
AdaMatch        & \underline{5.09}\tiny{±0.21}          & 5.13\tiny{±0.05}          & 4.36\tiny{±0.05}          & 37.08\tiny{±1.35}          & 26.66\tiny{±0.33}          & 21.99\tiny{±0.15}          & 6.14\tiny{±5.35}          & 2.13\tiny{±0.04}          & 2.02\tiny{±0.05}          & 19.95\tiny{±5.17}          & 8.59\tiny{±0.43}          & 6.01\tiny{±0.02}          &6.00   &5   &12.10 \\
SimMatch        & 5.38\tiny{±0.01}          & 5.36\tiny{±0.08}          & 4.41\tiny{±0.07}          & 39.32\tiny{±0.72}          & 26.21\tiny{±0.37}          & 21.50\tiny{±0.11}          & 7.60\tiny{±2.11}          & 2.48\tiny{±0.61}          & 2.05\tiny{±0.05}          & 16.98\tiny{±4.24}          & 8.27\tiny{±0.40}          & 5.74\tiny{±0.31}          &6.58   &6   &12.11 \\
SimMatchV2      & \textbf{4.90}\tiny{±0.16}          & 5.04\tiny{±0.09}          & 4.33\tiny{±0.16}          & 36.68\tiny{±0.86}          & 26.66\tiny{±0.38}          & 21.37\tiny{±0.20}          & 7.92\tiny{±2.80}          & 2.92\tiny{±0.81}          & 2.85\tiny{±0.91}          & 15.85\tiny{±2.62}          & 7.54\tiny{±0.81}          & 5.65\tiny{±0.26}          &5.25   &3   &11.81 \\
\cmidrule(r){1-1}\cmidrule(lr){2-4}\cmidrule(lr){5-7}\cmidrule(lr){8-10}\cmidrule(lr){11-13}\cmidrule(lr){14-14}\cmidrule(lr){15-15}\cmidrule(lr){16-16}
Super-SST (ours)& 9.59\tiny{±0.32}          & \underline{3.37}\tiny{±0.22}          & \underline{1.61}\tiny{±0.18}          & \underline{35.50}\tiny{±0.58}          & \underline{18.51}\tiny{±0.36}          & \underline{14.20}\tiny{±0.17}          & 29.41\tiny{±1.55}         & 4.17\tiny{±0.49}          & 3.10\tiny{±0.32}          & \underline{7.33}\tiny{±0.31}           & \underline{1.59}\tiny{±0.12}          & \underline{1.45}\tiny{±0.10}          &\underline{5.08}   &\underline{2}   &\underline{10.82} \\
Semi-SST (ours) & 6.35\tiny{±0.28}          & \textbf{2.42}\tiny{±0.13}          & \textbf{1.41}\tiny{±0.10}          & \textbf{31.39}\tiny{±0.47}          & \textbf{16.62}\tiny{±0.28}          & \textbf{13.50}\tiny{±0.14}          & 23.18\tiny{±1.38}         & 3.36\tiny{±0.17}          & 2.97\tiny{±0.11}          & \textbf{3.92}\tiny{±0.20}           & \textbf{1.40}\tiny{±0.10}          & \textbf{1.36}\tiny{±0.08}          &\textbf{3.58}   &\textbf{1}   &\textbf{8.99} \\
\bottomrule
\end{tabular}
}
\end{table}

The results presented in Table \ref{tab:classic-cv-results} clearly highlight the superior performance of Semi-SST and Super-SST compared to existing SSL methods across multiple benchmarks. Notably, Semi-SST achieves the lowest mean error rate of 8.99\%, ranking first overall, while Super-SST follows closely with a mean error rate of 10.82\%, securing the second position. Notably, despite the availability of 531,131 additional training images in SVHN and 100,000 unlabeled images in STL-10, our method achieves competitive performance without leveraging these extra resources, underscoring its data-efficient learning capability. To further assess the statistical significance of our results, we conducted a Friedman test, which yielded a p-value of $3.11\times10^{-25}$, indicating a statistically significant difference among the evaluated methods. The Friedman rankings for Semi-SST and Super-SST are 1st and 2nd, respectively, suggesting their consistent advantage across multiple benchmarks. These results provide strong statistical support for the effectiveness of our approach while reducing the likelihood that the observed improvements are due to random variation. Despite Semi-SST and Super-SST demonstrate strong overall performance, they exhibit relatively higher error rates on the SVHN dataset with only 40 labeled samples. This suggests that our methods, like other SSL techniques, may face challenges in scenarios with extremely limited labeled data. Future work could explore enhancements to further improve performance in such cases. To ensure consistency with the experimental configurations in this paper, we utilize a self-supervised pretrained ViT-Small from DINO, which differs from the architectures used in prior works such as SimMatchV2 \citep{zheng2023simmatchv2} and USB \citep{wang2022usb}. While this setup may result in a less direct comparison, it contributes valuable data points to the research community, offering insights into the performance of transformer-based architectures in SSL.

\subsection{Comparison with fully-supervised models}
\label{sec:comparison with fully-supervised models}

Table \ref{tab:comparison with fully-supervised methods} compares our SST methods with fully-supervised methods such as DeiT \citep{touvron2021training} and DeiT-III \citep{touvron2022deit} on ImageNet-1K. As shown, the fully-supervised DeiT-III-ViT-Huge model achieves 84.8\% Top-1 accuracy using 100\% labeled data. In comparison, our methods perform remarkably well even with significantly reduced labeled data. For instance, with only 10\% labeled data and ViT-Huge, Super-SST and Semi-SST achieve the same 84.8\% and higher 84.9\% Top-1 accuracy, respectively, suggesting a tenfold reduction in human annotation costs, and highlighting the effectiveness of our methods in utilizing unlabeled data. This demonstrates that our methods can be highly effective in applications where obtaining labeled data is challenging or costly, thereby significantly narrowing the performance disparity between semi-supervised and fully-supervised methods. Moreover, using huge-to-small distillation, Super-SST-ViT-Small achieves an impressive 80.3\% Top-1 accuracy on 10\% labeled data, outperforming the fully-supervised DeiT-ViT-Small's 79.8\% Top-1 accuracy on 100\% labeled data. 

\begin{table}
    \caption{Comparison with fully-supervised methods on 100\%, 10\%, and 1\% ImageNet-1K labeled data. $^\dagger$ indicates result at a resolution of 224$\times$224 for fair comparisons. Our methods significantly reduce the reliance on human annotation.}
    \label{tab:comparison with fully-supervised methods}
    \setlength{\tabcolsep}{18pt}
    \centering
    \small
    \begin{tabular*}{\linewidth}{llrccc}
        \toprule
        \multirow{2}{*}{Method}& \multirow{2}{*}{Architecture}& \multirow{2}{*}{Param}& \multicolumn{3}{c}{ImageNet-1K}\\
                               &                              &                       & 100\%& 10\%& 1\%\\\hline
        DeiT \citep{touvron2021training}& ViT-Small& 22M& 79.8& -& -\\
        Super-SST (ours)&                 ViT-Small& 22M& -& 78.3& 70.4\\
        Super-SST (ours, distilled)&      ViT-Small& 22M& -& 80.3& 76.9\\
        Semi-SST (ours)&                  ViT-Small& 22M& -& 78.6& 71.4\\\hline
        DeiT-III \citep{touvron2022deit}& ViT-Huge& 632M& 84.8$^\dagger$& -& -\\
        Super-SST (ours)&                 ViT-Huge& 632M& -& 84.8& 80.3\\
        Semi-SST (ours)&                  ViT-Huge& 632M& -& \textbf{84.9}& \textbf{80.7}\\
        \bottomrule
    \end{tabular*}
\end{table}

\subsection{Comparison of efficiency}
\label{sec:comparison of efficiency}

\begin{table}
\caption{Comparison of efficiency and performance on ImageNet-1K. Following FlexMatch and FreeMatch, we use 4 NVIDIA Tesla V100 GPUs for experiments on ImageNet-1K.}
\label{tab:comparisons of efficiency}
\centering
\begin{subtable}{1.0\textwidth}
    \centering
    \setlength\tabcolsep{18pt}
    \begin{tabular*}{\linewidth}{lcccccc}
    \toprule
    Method& Architecture& \makecell{Number of\\threshold updates}& \makecell{Runtime\\(sec./iter.)}& \makecell{ImageNet-1K\\100K}\\\hline
        FixMatch \citep{sohn2020fixmatch} & ResNet-50& 0& 0.4& 56.34\\
        FlexMatch \citep{zhang2021flexmatch} & ResNet-50& 1,048,576& 0.6& 58.15\\
        FreeMatch \citep{wang2023freematch} & ResNet-50& 1,048,576& 0.4& 59.43\\
        Super-SST (ours)& ResNet-50& \textbf{3}& \textbf{0.1}& \textbf{64.42}\\
    \bottomrule
    \end{tabular*}
    \caption{Comparison of runtime and Top-1 accuracy on ImageNet-1K with 100K randomly sampled labeled data (\ie, 100 labels per class). The experimental setting follows FreeMatch \citep{wang2023freematch} and FlexMatch \citep{zhang2021flexmatch} to maintain consistency. Following FlexMatch and FreeMatch, we train the ResNet-50 from scratch for fair comparisons. All other results are directly sourced from FreeMatch \citep{wang2023freematch}.}
    \label{tab:comparisons of runtime}
\end{subtable}\\
\begin{subtable}{1.0\textwidth}
    \centering
    \setlength\tabcolsep{17pt}
    \begin{tabular*}{\linewidth}{lccrrc}
        \toprule
        Method& Architecture& \makecell{Multi-crop\\strategy}& \makecell{GPU\\days}& \makecell{GPU\\memory}& \makecell{ImageNet-1K\\1\%}\\\hline
        CoMatch \citep{li2021comatch} & ResNet-50& \textcolor{gray!80}{\ding{55}}& 10.3& 81G& 61.1\\
        SimMatch \citep{zheng2022simmatch} & ResNet-50& \textcolor{gray!80}{\ding{55}}& \textbf{7.7}& 40G& 61.2\\
        PAWS \citep{assran2021semi} & ResNet-50& \ding{51}& 22.6& 1048G& 63.8\\
        SimMatchV2 \citep{zheng2023simmatchv2} & ResNet-50& \textcolor{gray!80}{\ding{55}}& 7.8& 42G& 63.7\\
        SimMatchV2 \citep{zheng2023simmatchv2} & ResNet-50& \ding{51}& 14.7& 83G& 69.9\\
        Super-SST (ours)& ViT-Small& \textcolor{gray!80}{\ding{55}}& 10.5& \textbf{36G}& 70.4\\
        Semi-SST (ours)&  ViT-Small& \textcolor{gray!80}{\ding{55}}& 28.6& 52G& \textbf{71.4}\\
        \bottomrule
    \end{tabular*}
    \caption{Comparison of GPU days, memory and Top-1 accuracy on ImageNet-1K with 1\% labels per class. The experimental setting follows SimMatchV2 \citep{zheng2023simmatchv2} to maintain consistency. All other results are directly sourced from SimMatchV2.}
    \label{tab:comparisons of GPU days and memory}
\end{subtable}
\end{table}

\begin{table}
\caption{Comparison of GPU Hours and Mean Error Rates on CIFAR-10, CIFAR-100, SVHN, and STL-10. The reported results of FreeMatch and SoftMatch are directly sourced from their papers. Mean error rates represent the average over three runs. Following FreeMatch and SoftMatch, we use a single V100 GPU for experiments on these datasets.}
\label{tab:comparisons of GPU hours}
\centering
\begin{subtable}{1.0\textwidth}
    \centering
    \setlength\tabcolsep{18pt}
    \begin{tabular*}{\linewidth}{lccccc}
    \toprule
    \multirow{2}{*}{Method}& \multicolumn{4}{c}{GPU hours}& \multirow{2}{*}{\makecell{Mean\\error rate}}\\
                                             & CIFAR-10& CIFAR-100& SVHN& STL-10& \\\hline
        SoftMatch \citep{chen2023softmatch} &  $\sim$72&  $\sim$168& $\sim$72& $\sim$168& 13.11\\
        FreeMatch \citep{wang2023freematch} &  $\sim$48&  $\sim$240& $\sim$48& $\sim$240& 12.51\\
        Super-SST (ours)&                       $\sim$3&    $\sim$3&  $\sim$7&   $\sim$2& 12.41\\
        Semi-SST (ours)&                        $\sim$4&    $\sim$4& $\sim$15&   $\sim$3& 10.31\\
    \bottomrule
    \end{tabular*}
\end{subtable}\\
\end{table}

Table \ref{tab:comparisons of efficiency} presents a comparison of efficiency and performance between our methods and other SSL models on ImageNet-1K, focusing on threshold updates, runtime, GPU utilization, and Top-1 accuracy.

\textbf{Comparison of threshold updates and runtime}. Table \ref{tab:comparisons of runtime} provides a detailed comparison of efficiency and performance between FixMatch, FlexMatch, FreeMatch, and Super-SST. Note that self-supervised pretraining is optional in our methods. Following the same protocol as FlexMatch and FreeMatch, we train the ResNet-50 from scratch on 100K ImageNet-1K labeled data (\ie, 100 labels per class) for fair comparisons. As shown, FixMatch uses a predefined constant threshold, resulting in zero threshold updates. Both FlexMatch and FreeMatch update the class-specific thresholds at each iteration, leading to over 1,000,000 updates, making them computationally intensive and costly. Super-SST trains the ResNet-50 for 3 cycles and updates the class-specific thresholds only at each training cycle, resulting in just 3 updates. This vast difference in the number of updates makes Super-SST much more efficient in terms of time and computational costs. Additionally, Super-SST updates thresholds using a well-trained high-performance model, reducing erroneous predictions and confirmation bias. In contrast, FreeMatch and FlexMatch update thresholds using models that are still in training, leading to more inaccurate predictions during the early stages and amplifying confirmation bias, ultimately resulting in poorer performance. Super-SST achieves the fastest runtime of 0.1 seconds per iteration, compared to 0.4 seconds for FixMatch and FreeMatch, and 0.6 seconds for FlexMatch. This makes Super-SST the most efficient method with faster runtime and fewer threshold updates among those compared. Furthermore, Super-SST not only excels in efficiency but also achieves the highest Top-1 accuracy of 64.42\%, significantly outperforming all counterparts by a substantial margin. 

\textbf{Comparison of GPU days and memory}. Table \ref{tab:comparisons of GPU days and memory} compares GPU days, GPU memory and Top-1 accuracy between CoMatch, SimMatch, PAWS, SimMatchV2, Super-SST, and Semi-SST on 1\% ImageNet-1K labeled data. As shown, PAWS and SimMatchV2 use the multi-crop strategy, processing multiple cropped versions of each image, leading to better model performance but significantly higher computational demands and longer training time. For example, the training time of SimMatchV2 nearly doubles (from 7.8 to 14.7 days) when multi-crops are included, and its Top-1 accuracy significantly drops by 6.2\% when multi-crops are excluded. Super-SST stands out for its excellent balance of efficiency and performance, achieving the second best Top-1 accuracy of 70.4\% with the lowest GPU memory usage and relatively moderate GPU days, making it highly suitable for practical applications where both accuracy and efficiency are crucial. Semi-SST achieves the highest accuracy of 71.4\% at the cost of high GPU days and moderate GPU memory usage, making it suitable for scenarios where maximum accuracy is required and computational resources are less of a constraint. Our methods outperform others in accuracy while maintaining reasonable resource usage, marking significant advancements in the SSL field.

\textbf{Comparison of GPU hours}. Table \ref{tab:comparisons of GPU hours} provides a comprehensive comparison of the computational efficiency and mean error rates of FreeMatch, SoftMatch, Super-SST, and Semi-SST across multiple benchmark datasets, including CIFAR-10, CIFAR-100, SVHN, and STL-10. When computing mean error rates, we reuse the experimental results from Table \ref{tab:classic-cv-results}. It is important to note that FreeMatch and SoftMatch employ a slightly different set of benchmark configurations compared to our study, as well as SimMatchV2 and USB. Specifically, FreeMatch reports results for CIFAR-10 with 10 labels but does not provide results for STL-10 with 250 labels. Meanwhile, SoftMatch does not report results for SVHN and STL-10 with 250 labels. To align with USB and SimMatchV2, our study excludes CIFAR-10 with 10 labels but includes SVHN and STL-10 with 250 labels. To ensure a fair and consistent comparison, we consider the intersection of these benchmark configurations, ultimately including 10 settings: CIFAR-10 (with 40, 250, and 4000 labels), CIFAR-100 (with 400, 2500, and 10000 labels), SVHN (with 40 and 1000 labels), and STL-10 (with 40 and 1000 labels). As shown in Table \ref{tab:comparisons of GPU hours}, Super-SST and Semi-SST significantly reduce computational costs while maintaining superior performance. On CIFAR-100, FreeMatch requires approximately 240 GPU hours, whereas Super-SST and Semi-SST achieve better performance with about 3 and 4 GPU hours, respectively, representing speedups of 80$\times$ and 60$\times$. Similarly, on STL-10, FreeMatch consumes about 240 GPU hours, while Super-SST and Semi-SST require approximately 2 and 3 GPU hours, respectively, reducing the computational cost by 120$\times$ and 80$\times$. A similar trend is observed on CIFAR-10 and SVHN, where Super-SST and Semi-SST consistently require an order of magnitude fewer GPU hours than FreeMatch and SoftMatch. Beyond efficiency, Semi-SST achieves the best overall performance, attaining a mean error rate of 10.31, which is lower than both SoftMatch (13.11), FreeMatch (12.51) and Super-SST (12.41). Notably, this improvement is achieved with drastically fewer GPU hours, further demonstrating the effectiveness of our approach in mitigating confirmation bias while maintaining a lightweight computational footprint. The results highlight the efficiency and effectiveness of our proposed SST methods. By significantly reducing training time while maintaining or surpassing state-of-the-art performance, our approach offers a compelling alternative to existing semi-supervised learning methods. These findings underscore the potential of SST as a highly cost-effective solution for large-scale SSL applications, particularly in scenarios where computational resources are constrained. Note that we use ViT-Small architecture, while FreeMatch and SoftMatch employ Wide ResNet (WRN) architecture \citep{zagoruyko2016wide}, which may result in a less direct comparison.

\subsection{Class imbalance and label noise}
\label{sec:class imbalance and label noise}

\begin{table}
    \caption{Comparison with supervised baseline on Clothing-1M dataset. The performance gap to the baseline is indicated in bracket. The result of supervised baseline is implemented by us.}
    \label{tab:comparison on clothing-1m}
    \setlength{\tabcolsep}{13pt}
    \centering
    \begin{tabular*}{\linewidth}{llllll}
        \toprule
        Dataset& Architecture& Parameter& Method& 5\%& 100\%\\\hline
        \multirow{2}{*}{Clothing-1M} & \multirow{2}{*}{ViT-Small}& \multirow{2}{*}{22M}& Supervised& 71.28& 78.99\\
                                     &                           &                     & Super-SST (ours)& 75.69 (\textcolor{hscolor}{+4.41})& - \\
        \bottomrule
    \end{tabular*}
\end{table}

Clothing-1M \citep{xiao2015learning} is a large-scale dataset containing approximately 1 million clothing images collected from online shopping websites. The dataset exhibits both severe class imbalance and label noise, making it a particularly challenging benchmark for robust learning algorithms. The class distribution is highly skewed, with some common clothing categories (e.g., T-shirts, jackets) being vastly overrepresented, while others (e.g., swimwear, scarves) have significantly fewer samples. This imbalance, combined with noisy annotations, increases the difficulty of model training, as standard learning methods may be biased toward high-frequency categories and struggle to generalize to underrepresented classes. We randomly sample 5\% samples from the clean-labeled training images, with the remaining images combined with 1M noisy samples as the unlabeled data. As shown in Table \ref{tab:comparison on clothing-1m}, Super-SST achieves 75.69\% top-1 accuracy with only 5\% of Clothing-1M clean-labeled data, surpassing the supervised baseline by 4.41\% and demonstrating strong robustness to both class imbalance and label noise. Notably, this performance is close to the fully supervised upper bound of 78.99\%, despite relying on significantly fewer labeled samples, further highlighting its effectiveness in real-world imbalanced scenarios.

Similarly, iNaturalist \citep{inaturalist-2019-fgvc6} is a large-scale, real-world dataset that exhibits extreme class imbalance, as the number of images per class follows a long-tailed distribution. A few dominant species have thousands of images, while many rare species are represented by only a handful of samples. This highly skewed distribution poses significant challenges for learning algorithms, as models trained on iNaturalist tend to be biased toward majority classes, struggling to generalize to underrepresented species. Consequently, iNaturalist serves as a crucial benchmark for evaluating techniques designed to address long-tailed recognition, class-balanced learning, and data reweighting strategies. As shown in Table \ref{tab:comparison with baselines}, our method achieves significant improvements under extreme class imbalance conditions. With 1\% / 10\% labeled iNaturalist data and ViT-Base, Super-SST outperforms the supervised baseline by 3.0\% / 7.8\%, while Semi-SST achieves a 1.1\% / 0.4\% relative improvement over the semi-supervised baseline. These results demonstrate that our method effectively mitigates the impact of class imbalance and enhances model generalization under highly skewed distributions, further reinforcing its applicability to real-world long-tailed learning challenges.

\subsection{Ablation study}

\begin{figure}
    \centering
    \begin{subfigure}[c]{0.42\textwidth}
        \centering
        \includegraphics[width=\textwidth]{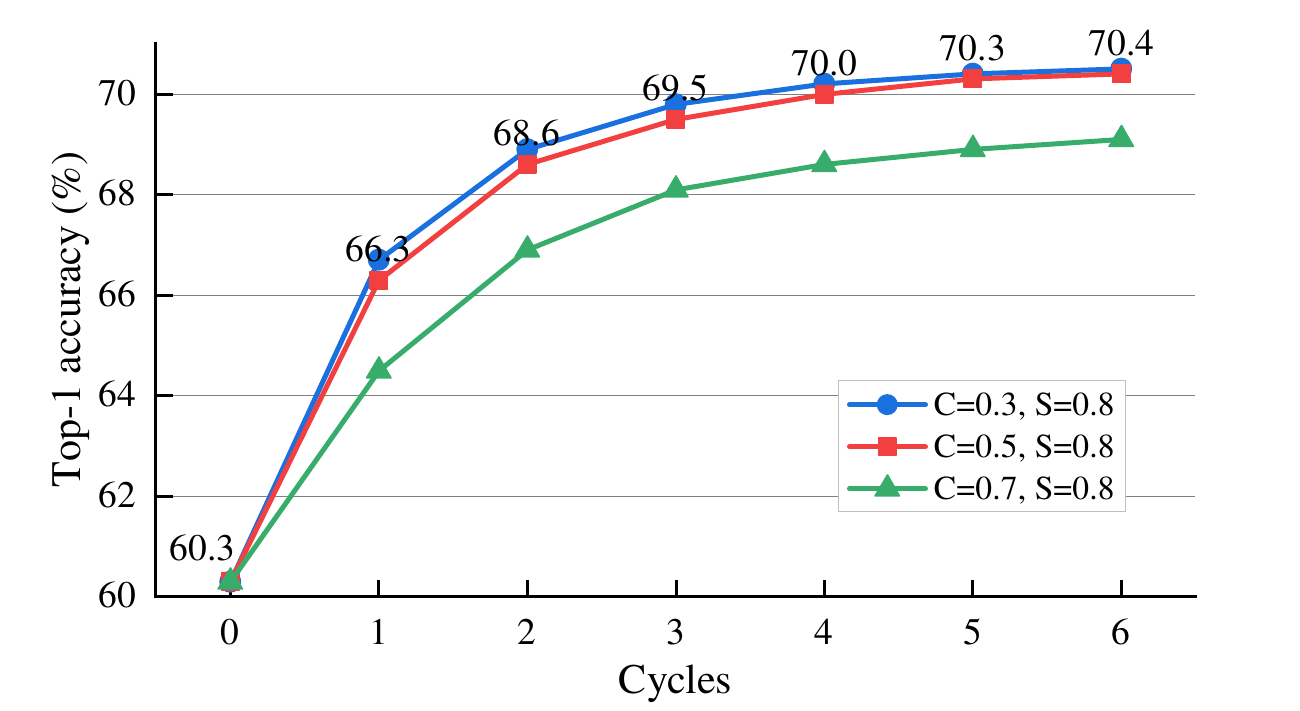}
        \caption{Influence of cutoff value $C$}
        \label{fig:influence_of_c}
    \end{subfigure}
    \hspace{15pt}
    \begin{subfigure}[c]{0.42\textwidth}
        \centering
        \includegraphics[width=\textwidth]{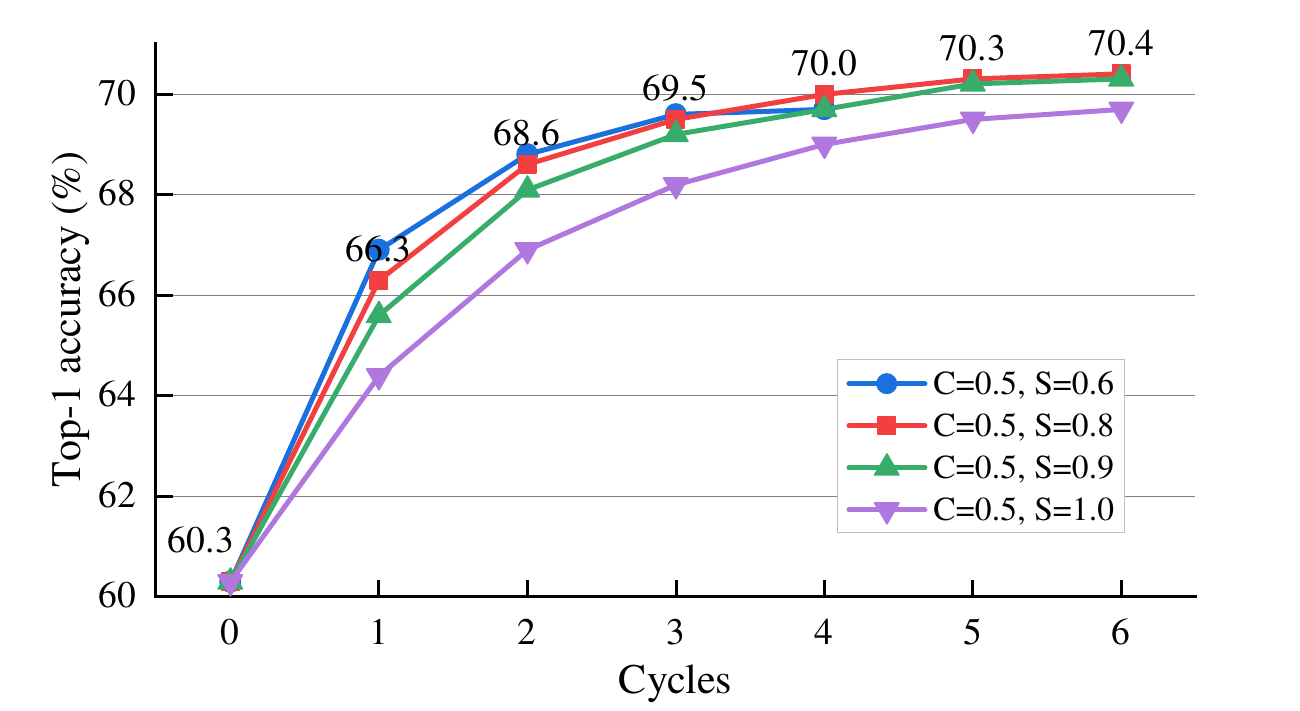}
        \caption{Influence of scaling factor $S$}
        \label{fig:influence_of_s}
    \end{subfigure}
    \caption{Influence of cutoff value ($C$) and scaling factor ($S$) in SAT. All experiments are conducted with the same starting checkpoint (ViT-Small) that is supervised finetuned on 1\% ImageNet-1K labels, achieving 60.3\% accuracy.}
    \label{fig:influence_of_c_s}
\end{figure}

Figure \ref{fig:influence_of_c_s} explores the impact of different cutoff values $C$ and scaling factors $S$ on Top-1 accuracy over multiple training cycles. Each subplot focuses on a specific variable while keeping the other constant to observe their individual effects on performance. All experiments are conducted with the same starting checkpoint (\ie, ViT-Small) that is supervised finetuned on 1\% ImageNet-1K labeled data, achieving 60.3\% Top-1 accuracy. As shown in Figure \ref{fig:influence_of_c}, lower cutoff values (C=0.3 and 0.5) lead to faster and higher improvements in accuracy. Higher cutoff value (C=0.7) underperforms compared to lower cutoff values, indicating that a high cutoff value might hinder model performance. As shown in Figure \ref{fig:influence_of_s}, lower scaling factor (S=0.6) leads to faster improvements in accuracy but quickly stabilizes at a lower level, indicating insufficient improvement over multiple training cycles. A higher scaling factor (S=1.0) results in slower improvements and lower peak accuracy. Moderate scaling factors (S=0.8 and 0.9) provide better performance. Based on these observations, we set $C$ to 0.5 and $S$ to 0.8 for ViT-Small and 0.9 for ViT-Huge on ImageNet-1K. Detailed hyper-parameter settings are provided in Table \ref{tab:settings}.

\begin{figure}
    \centering
    \begin{subfigure}[c]{0.42\textwidth}
        \centering
        \includegraphics[width=\textwidth]{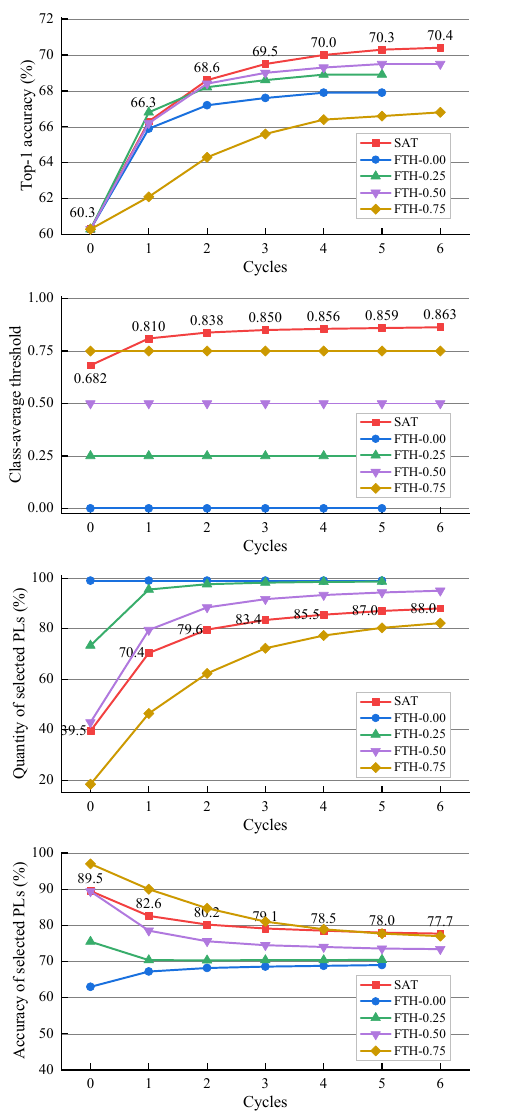}
        \caption{Super-SST}
        \label{fig:sat_vs_fth_super}
    \end{subfigure}
    \hspace{15pt}
    \begin{subfigure}[c]{0.42\textwidth}
        \centering
        \includegraphics[width=\textwidth]{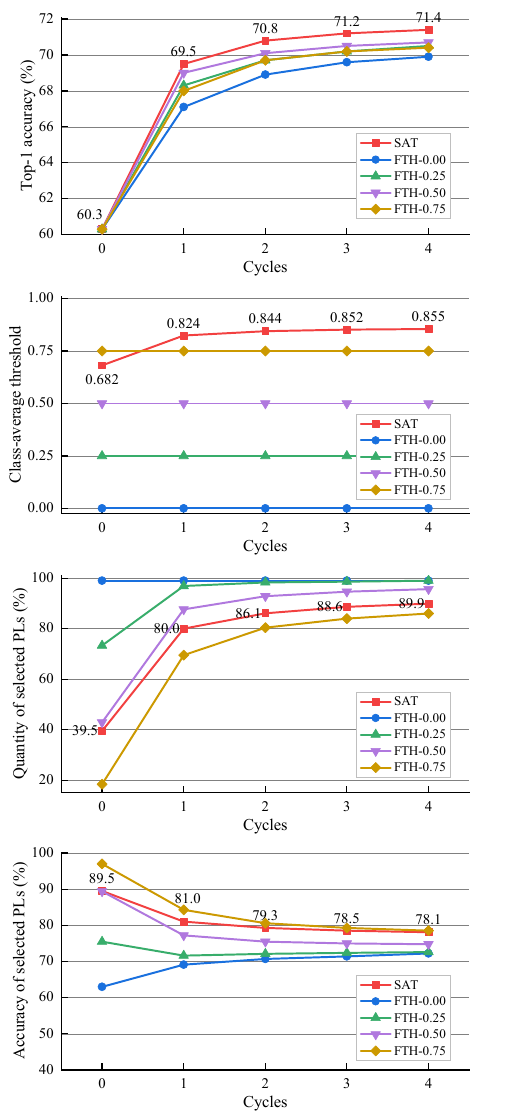}
        \caption{Semi-SST}
        \label{fig:sat_vs_fth_semi}
    \end{subfigure}
    \caption{Self-Adaptive Thresholding (SAT) vs. Fixed Thresholding (FTH). In this figure, we set the cutoff value $C$=0.5 and scaling factor $S$=0.8 for SAT. For FTH, we set the constant thresholds to 0.00, 0.25, 0.50, and 0.75, respectively.}
    \label{fig:sat_vs_fth}
\end{figure}

Figure \ref{fig:sat_vs_fth} presents a comprehensive comparison between Self-Adaptive Thresholding (SAT) and various Fixed Thresholding (FTH) settings for both Super-SST and Semi-SST. The metrics analyzed include Top-1 accuracy, class-average threshold, quantity and accuracy of selected pseudo-labels (PLs) over multiple training cycles. All experiments are conducted with the same starting checkpoint (\ie, ViT-Small) that is supervised finetuned on 1\% ImageNet-1K labeled data, achieving 60.3\% Top-1 accuracy. 

\textbf{Top-1 accuracy}. In Super-SST / Semi-SST, SAT starts with a Top-1 accuracy of 60.3\% and shows a rapid increase to 66.3\% / 69.5\% in Cycle 1, peaking at 70.4\% / 71.4\% by Cycle 6 / 4. While the FTH settings exhibit similar upward trends, their final performance is lower than that of SAT. 

\textbf{Class-average threshold}. In Super-SST and Semi-SST, SAT starts with a class-average threshold of 0.682, gradually increasing to 0.863 and 0.855, respectively. FTH methods maintain constant thresholds (\ie, 0.00, 0.25, 0.50, and 0.75) due to their fixed nature, with no variation across cycles. The dynamic nature of SAT allows it to accurately and adaptively adjust class-specific thresholds according to the model’s performance, providing a more robust and responsive learning process compared to the static FTH methods. 

\textbf{Quantity and accuracy of selected PLs}. The balance between high quality and sufficient quantity of selected PLs is crucial. In Super-SST / Semi-SST, SAT starts with 39.5\% selected PLs and gradually increases to 88.0\% / 89.9\%. Additionally, the accuracy of the selected PLs begins at 89.5\% and gradually decreases to 77.7\% / 78.1\%, possibly due to the incorporation of more noisy labels. Our SAT mechanism stands out for its excellent balance of high quality and sufficient quantity of selected PLs. Compared to SAT, FTH methods struggle to balance the high quality and sufficient quantity of selected PLs. For instance, FTH-0.00 uses all PLs without discrimination, leading to the lowest accuracy of PLs. FTH-0.75 maintains similar accuracy of PLs to SAT but often selects fewer PLs than SAT. 

In summary, Figure \ref{fig:sat_vs_fth}  highlights the clear superiority of SAT over FTH methods in both Super-SST and Semi-SST frameworks. SAT's dynamic thresholding mechanism allows for consistent improvements in top-1 accuracy and effective balancing of the high quality and sufficient quantity of selected PLs, making it a superior choice for SSL tasks. The static nature of FTH methods means they cannot respond to the changing dynamics of training, resulting in either overly conservative or overly liberal PLs selection, leading to limited improvements in model performance.

\section{Related work}
\label{sec:related work}

\subsection{Thresholding techniques}

Our work is closely related to SSL methods that utilize thresholding techniques \citep{cai2022semi, cai2021exponential, xie2020self, zhang2021flexmatch, wang2023freematch, sohn2020fixmatch, xu2021dash, berthelot2021adamatch, guo2022class, lin2019towards, chen2020self, kumar2020understanding, chen2023softmatch}. Methods such as Semi-ViT \citep{cai2022semi}, Noisy Student \citep{xie2020self}, UDA \citep{xie2020unsupervised}, and FixMatch \citep{sohn2020fixmatch} rely on fixed thresholding strategy to select unlabeled samples with predicted confidence scores above a predefined constant. Dash \citep{xu2021dash} employs a gradually increasing threshold to incorporate more unlabeled data into early-stage training. Adsh \citep{guo2022class} derives adaptive thresholds for imbalanced SSL by optimizing pseudo-label numbers per class. AdaMatch \citep{berthelot2021adamatch} uses an adaptively estimated relative threshold based on the EMA of labeled data confidence. FlexMatch \citep{zhang2021flexmatch} uses a higher constant as dataset-specific global threshold and adjusts it according to the number of confident unlabeled samples predicted to each class. FreeMatch \citep{wang2023freematch} estimates a dataset-specific global threshold through the EMA of the unlabeled data confidence and further adjusts it based on the EMA expectation of each class's predictions. SoftMatch \citep{chen2023softmatch} introduces a novel approach to semi-supervised learning by addressing the quantity-quality trade-off in pseudo-labeling, balancing the use of abundant low-confidence samples with high-confidence ones. By incorporating a soft thresholding mechanism, SoftMatch effectively improves label efficiency and enhances model performance across various SSL benchmarks. USB \citep{wang2022usb} is a unified and challenging SSL benchmark with 15 tasks across computer vision, natural language processing, and audio for fair and consistent evaluations. USB provides the implementation of multiple SSL algorithms, including FixMatch, FlexMatch, SimMatch, SoftMatch, and FreeMatch.

Our SAT mechanism distinguishes itself from existing dynamic thresholding techniques in three main aspects: generation mechanism, number of threshold updates, and performance improvement. SAT is a confidence-based method that initially uses a cutoff value $C$ to discard the lower and unreliable probabilities for each class. It then calculates the average of the remaining higher and reliable probabilities for each class and multiplies this by a scaling factor $S$ to obtain the class-specific thresholds. Unlike FlexMatch, our SAT does not rely on the number of unlabeled samples predicted to each class, nor does it use a dataset-specific global threshold. Unlike FreeMatch, SAT employs a cutoff mechanism that FreeMatch lacks. Additionally, SAT does not use EMA and global threshold as in FreeMatch. The most significant difference between SAT and FlexMatch or FreeMatch lies in the number of threshold updates. FlexMatch and FreeMatch update the thresholds at each iteration using a model that is still in training, leading to over 1,000,000 updates. This incurs extremely high computational and time costs and tends to amplify confirmation bias during the early stage of training, thereby leading to poorer model performance. In contrast, SAT updates the thresholds once per training cycle using a well-trained high-performance model, resulting in a single-digit number of updates and negligible computational and time costs. Detailed theoretical analysis and extensive experimental results strongly confirm the innovation, uniqueness, and superiority of SAT. These advantages enable SST to significantly outperform previous counterparts in both performance and efficiency, establishing it as a milestone in the field of SSL.

\subsection{Self-training and pseudo-labeling}

Self-training and pseudo-labeling are closely related methods in SSL, often used together. Self-training is a comprehensive framework that iteratively uses pseudo-labeling along with other strategies to enhance the performance of model. Pseudo-labeling is a specific technique that generates pseudo-labels for unlabeled data, making it an integral part of self-training frameworks. The principles of self-training and pseudo-labeling, originating from early research \citep{scudder1965probability, mclachlan1975iterative, yarowsky1995unsupervised, riloff1996automatically, riloff2003learning, lee2013pseudo}, have been revitalized in recent SSL studies \citep{shi2018transductive, iscen2019label, arazo2020pseudo, sohn2020simple, xie2020self, xie2020unsupervised, sohn2020fixmatch, pham2021meta, cai2021exponential, liu2021unbiased, zhang2021flexmatch, wei2021crest, chen2022debiased, wang2023freematch}. Pseudo-Labeling \citep{lee2013pseudo} utilizes hard pseudo-labels derived from model predictions to guide the self-training process on unlabeled data. Noisy Student \citep{xie2020self} incorporates self-training with noise injection to enhance model performance. CBST \citep{zou2018unsupervised} addresses the challenge of class imbalance in semantic segmentation by generating pseudo-labels with balanced class distribution. CReST \citep{wei2021crest} uses class-rebalancing technique that adjusts the number of pseudo-labeled samples per class based on class-specific sampling rates. Debiased Self-Training \citep{chen2022debiased} aims to mitigate training bias in self-training by using independent heads for generating and utilizing pseudo-labels. ST++ \citep{yang2022st++} extends traditional self-training methods by exploiting progressively refined pseudo-labels in a curriculum learning manner \citep{bengio2009curriculum}. UDA \citep{xie2020unsupervised} enhances SSL models by sharpening model predictions instead of converting them into hard pseudo-labels. FixMatch \citep{sohn2020fixmatch} converts weakly-augmented predictions into hard pseudo-labels for strongly-augmented data. Label Propagation \citep{iscen2019label} employs a transductive approach, predicting the entire dataset to generate pseudo-labels. SimMatch \citep{zheng2022simmatch} leverages a labeled memory buffer to propagate semantic and instance pseudo-labels.

Our SST method stands out from traditional self-training and pseudo-labeling methods through several key innovations. The primary distinction is the combination of self-training with our innovative SAT mechanism, which accurately and adaptively adjusts class-specific thresholds according to the learning progress of the model. This mechanism ensures both high quality and sufficient quantity of selected pseudo-labels, reducing the risks of inaccuracies and confirmation bias, thereby enhancing overall model performance. Additionally, SST framework demonstrates significant flexibility as it can be integrated not only with supervised algorithms but also with semi-supervised algorithms like EMA-Teacher \citep{cai2021exponential, cai2022semi}. This flexibility also distinguishes it from traditional self-training methods.

\section{Limitation}
\label{sec:limitation}

Despite the strong overall performance demonstrated by Semi-SST and Super-SST, they exhibit relatively higher error rates under extremely limited labeled data scenarios (e.g., SVHN with only 40 labeled samples). This suggests that our methods may face challenges in ultra-low-label scenarios. The limitation aligns with broader challenges in SSL, where extreme label scarcity amplifies the sensitivity to pseudo-label noise and model initialization. Future work could explore strategies to mitigate this issue. Furthermore, our method is specifically designed for computer vision tasks and is not currently configured for natural language processing (NLP) benchmarks. Although experiments on text data are of interest, adapting our framework to NLP tasks would require significant architectural modifications. Future research could investigate ways to generalize our approach to broader modalities, including NLP and multi-modal learning. Additionally, while we have conducted experiments on the Clothing-1M \citep{xiao2015learning} dataset and provided corresponding results, our investigation into noisy data, class imbalance, and domain shift scenarios remains limited. Due to constraints on computational resources, our analysis of these challenges is not as comprehensive as desired. Future iterations of this work will aim to systematically explore these challenges, thereby offering deeper insights into SSL performance under more realistic and diverse conditions.

\section{Conclusion}
\label{sec:conclusion}
In this paper, we introduce an innovative Self-training with Self-adaptive Thresholding (SST) framework, along with its two variations: Super-SST and Semi-SST, significantly advancing the SSL field. The core innovation of our framework is the Self-Adaptive Thresholding (SAT) mechanism, which accurately and adaptively adjusts class-specific thresholds according to the learning progress of the model. SAT mechanism ensures both high quality and sufficient quantity of selected pseudo-labeled data, mitigating the risks of inaccurate pseudo-labels and confirmation bias, thereby enhancing overall model performance. Extensive experiments and results strongly confirm the effectiveness, efficiency, generalization, and scalability of SST framework across various architectures and datasets. Notably, Semi-SST-ViT-Huge achieves the best results on competitive ImageNet-1K SSL benchmarks, with 80.7\% / 84.9\% Top-1 accuracy using only 1\% / 10\% labeled data. Compared to the fully-supervised DeiT-III-ViT-Huge, which achieves 84.8\% Top-1 accuracy using 100\% labeled data, our method demonstrates superior performance using only 10\% labeled data. This indicates a tenfold reduction in human annotation costs, significantly narrowing the gap between semi-supervised and fully-supervised learning. Moreover, SST achieves SOTA performance across various architectures and datasets while demonstrating remarkable efficiency, resulting in an exceptional cost-performance ratio. These advancements lay the groundwork for further innovations in SSL and practical applications where acquiring labeled data is difficult or expensive. In future research, we intend to extend our methods to more practical applications and enhance our thresholding techniques and model optimization strategies. We anticipate that this work will motivate future studies and contribute valuable insights for the broader community. 

\textbf{Broader impacts}. Our methods have significant potential applications in fields where obtaining sufficient high-quality labeled data is difficult or unaffordable, such as medical research and defect detection. However, it is important to acknowledge that many people rely on providing annotation services for their livelihood, and reducing the need for these services may lead to decreased income or even unemployment for those individuals. Balancing the benefits of advanced SSL techniques with their socioeconomic impacts will be crucial as these technologies continue to develop.

\section{Acknowledgments}

We thank all the reviewers for their valuable suggestions and comments. Additionally, we express our sincere gratitude to the authors of Semi-ViT for generously sharing their code with the community. This work is supported by the National Natural Science Foundation of China (No. U21B2009, No. U19B2020, No. 12201057).

\appendix
\section{Appendix}

Our training primarily follows the hyper-parameter settings in Semi-ViT \citep{cai2022semi}, with some minor modifications. The hyper-parameter settings for supervised initialization, Super-SST, and Semi-SST on ImageNet-1K are provided in Table \ref{tab:settings}. The experimental settings for other datasets are almost identical to those used for ImageNet-1K, with only minor differences, and will be made available on request.

\begin{table}
    \caption{Hyper-parameter settings for supervised initialization, Super-SST, and Semi-SST on ImageNet-1K. S and H denote ViT-Small and ViT-Huge, respectively.}
    \label{tab:settings}
    \setlength{\tabcolsep}{9pt}
    \centering
    \footnotesize
    \begin{tabular*}{\linewidth}{l|cccc}
        
        \bottomrule
        \multirow{2}{*}{config}& \multicolumn{2}{c}{Supervised initialization}&   Super-SST&    Semi-SST\\
                               &                    1\%&                  10\%& 1\% \& 10\%& 1\% \& 10\%\\\hline
        base learning rate&     1.0e-4 (S), 1.0e-4 (H)& 5.0e-5 (S), 1.0e-3 (H)& 5.0e-5 (S), 1.0e-3 (H)& 2.5e-4 (S), 2.5e-3 (H)\\
        batch size&             512 (S), 256 (H)&       1024 (S), 256 (H)& 1024 (S), 256 (H)& 512 (S), 128 (H)\\
        layer-wise lr decay&    0.55 (S), 0.90 (H)&     0.65 (S), 0.80 (H)& 0.65 (S), 0.75 (H)& 0.65 (S), 0.75 (H)\\
        weight decay&           0.00 (S), 0.30 (H)&     0.05 (S), 0.50 (H)& 0.05 (S), 0.05 (H)& 0.05 (S), 0.05 (H)\\
        drop path&              0.05 (S), 0.05 (H)&     0.10 (S), 0.10 (H)& 0.10 (S), 0.20 (H)& 0.10 (S), 0.10 (H)\\
        mixup&                  0.40 (S), 0.30 (H)&     0.60 (S), 0.70 (H)& 0.60 (S), 0.80 (H)& 0.60 (S), 0.80 (H)\\
        cutmix&                 0.10 (S), 0.00 (H)&     1.00 (S), 1.00 (H)& 1.00 (S), 1.00 (H)& 1.00 (S), 1.00 (H)\\
        random erasing&         0.05 (S), 0.00 (H)&     0.25 (S), 0.30 (H)& 0.25 (S), 0.25 (H)& 0.25 (S), 0.25 (H)\\
        cutoff value&                            -&                      -& 0.50 (S), 0.50 (H)& 0.50 (S), 0.50 (H)\\
        scaling factor&                          -&                      -& 0.80 (S), 0.90 (H)& 0.80 (S), 0.90 (H)\\
        optimizer&              \multicolumn{4}{c}{AdamW}\\
        training epochs&        \multicolumn{4}{c}{100 (S), 50 (H)}\\
        warmup epochs&          \multicolumn{4}{c}{5}\\
        optimizer momentum&     \multicolumn{4}{c}{$\beta_1$, $\beta_2$=0.9, 0.999}\\
        learning rate schedule& \multicolumn{4}{c}{cosine decay}\\
        label smoothing&        \multicolumn{4}{c}{0.10}\\
        \toprule
    \end{tabular*}
\end{table}


\bibliographystyle{model1-num-names}

\bibliography{cas-refs}

\begin{thebibliography}{85}
\expandafter\ifx\csname natexlab\endcsname\relax\def\natexlab#1{#1}\fi
\providecommand{\url}[1]{\texttt{#1}}
\providecommand{\href}[2]{#2}
\providecommand{\path}[1]{#1}
\providecommand{\DOIprefix}{doi:}
\providecommand{\ArXivprefix}{arXiv:}
\providecommand{\URLprefix}{URL: }
\providecommand{\Pubmedprefix}{pmid:}
\providecommand{\doi}[1]{\href{http://dx.doi.org/#1}{\path{#1}}}
\providecommand{\Pubmed}[1]{\href{pmid:#1}{\path{#1}}}
\providecommand{\bibinfo}[2]{#2}
\ifx\xfnm\relax \def\xfnm[#1]{\unskip,\space#1}\fi
\bibitem[{Russakovsky et~al.(2015)Russakovsky, Deng, Su, Krause, Satheesh, Ma, Huang, Karpathy, Khosla, Bernstein et~al.}]{russakovsky2015imagenet}
\bibinfo{author}{O.~Russakovsky}, \bibinfo{author}{J.~Deng}, \bibinfo{author}{H.~Su}, \bibinfo{author}{J.~Krause}, \bibinfo{author}{S.~Satheesh}, \bibinfo{author}{S.~Ma}, \bibinfo{author}{Z.~Huang}, \bibinfo{author}{A.~Karpathy}, \bibinfo{author}{A.~Khosla}, \bibinfo{author}{M.~Bernstein}, et~al.,
\newblock \bibinfo{title}{Imagenet large scale visual recognition challenge},
\newblock \bibinfo{journal}{International journal of computer vision} \bibinfo{volume}{115} (\bibinfo{year}{2015}) \bibinfo{pages}{211--252}.
\bibitem[{Deng et~al.(2009)Deng, Dong, Socher, Li, Li, and Fei-Fei}]{deng2009imagenet}
\bibinfo{author}{J.~Deng}, \bibinfo{author}{W.~Dong}, \bibinfo{author}{R.~Socher}, \bibinfo{author}{L.-J. Li}, \bibinfo{author}{K.~Li}, \bibinfo{author}{L.~Fei-Fei},
\newblock \bibinfo{title}{Imagenet: A large-scale hierarchical image database},
\newblock in: \bibinfo{booktitle}{2009 IEEE conference on computer vision and pattern recognition}, \bibinfo{organization}{Ieee}, \bibinfo{year}{2009}, pp. \bibinfo{pages}{248--255}.
\bibitem[{Lin et~al.(2014)Lin, Maire, Belongie, Hays, Perona, Ramanan, Doll{\'a}r, and Zitnick}]{lin2014microsoft}
\bibinfo{author}{T.-Y. Lin}, \bibinfo{author}{M.~Maire}, \bibinfo{author}{S.~Belongie}, \bibinfo{author}{J.~Hays}, \bibinfo{author}{P.~Perona}, \bibinfo{author}{D.~Ramanan}, \bibinfo{author}{P.~Doll{\'a}r}, \bibinfo{author}{C.~L. Zitnick},
\newblock \bibinfo{title}{Microsoft coco: Common objects in context},
\newblock in: \bibinfo{booktitle}{Computer Vision--ECCV 2014: 13th European Conference, Zurich, Switzerland, September 6-12, 2014, Proceedings, Part V 13}, \bibinfo{organization}{Springer}, \bibinfo{year}{2014}, pp. \bibinfo{pages}{740--755}.
\bibitem[{Zhu(2005)}]{zhu2005semi}
\bibinfo{author}{X.~J. Zhu},
\newblock \bibinfo{title}{Semi-supervised learning literature survey}  (\bibinfo{year}{2005}).
\bibitem[{Zhu and Goldberg(2022)}]{zhu2022introduction}
\bibinfo{author}{X.~Zhu}, \bibinfo{author}{A.~B. Goldberg}, \bibinfo{title}{Introduction to semi-supervised learning}, \bibinfo{year}{2022}.
\bibitem[{Sohn et~al.(2020)Sohn, Berthelot, Carlini, Zhang, Zhang, Raffel, Cubuk, Kurakin, and Li}]{sohn2020fixmatch}
\bibinfo{author}{K.~Sohn}, \bibinfo{author}{D.~Berthelot}, \bibinfo{author}{N.~Carlini}, \bibinfo{author}{Z.~Zhang}, \bibinfo{author}{H.~Zhang}, \bibinfo{author}{C.~A. Raffel}, \bibinfo{author}{E.~D. Cubuk}, \bibinfo{author}{A.~Kurakin}, \bibinfo{author}{C.-L. Li},
\newblock \bibinfo{title}{Fixmatch: Simplifying semi-supervised learning with consistency and confidence},
\newblock \bibinfo{journal}{Advances in neural information processing systems} \bibinfo{volume}{33} (\bibinfo{year}{2020}) \bibinfo{pages}{596--608}.
\bibitem[{Van~Engelen and Hoos(2020)}]{van2020survey}
\bibinfo{author}{J.~E. Van~Engelen}, \bibinfo{author}{H.~H. Hoos},
\newblock \bibinfo{title}{A survey on semi-supervised learning},
\newblock \bibinfo{journal}{Machine learning} \bibinfo{volume}{109} (\bibinfo{year}{2020}) \bibinfo{pages}{373--440}.
\bibitem[{Laine and Aila(2016)}]{laine2016temporal}
\bibinfo{author}{S.~Laine}, \bibinfo{author}{T.~Aila},
\newblock \bibinfo{title}{Temporal ensembling for semi-supervised learning},
\newblock \bibinfo{journal}{arXiv preprint arXiv:1610.02242}  (\bibinfo{year}{2016}).
\bibitem[{Tarvainen and Valpola(2017)}]{tarvainen2017mean}
\bibinfo{author}{A.~Tarvainen}, \bibinfo{author}{H.~Valpola},
\newblock \bibinfo{title}{Mean teachers are better role models: Weight-averaged consistency targets improve semi-supervised deep learning results},
\newblock \bibinfo{journal}{Advances in neural information processing systems} \bibinfo{volume}{30} (\bibinfo{year}{2017}).
\bibitem[{Miyato et~al.(2018)Miyato, Maeda, Koyama, and Ishii}]{miyato2018virtual}
\bibinfo{author}{T.~Miyato}, \bibinfo{author}{S.-i. Maeda}, \bibinfo{author}{M.~Koyama}, \bibinfo{author}{S.~Ishii},
\newblock \bibinfo{title}{Virtual adversarial training: a regularization method for supervised and semi-supervised learning},
\newblock \bibinfo{journal}{IEEE transactions on pattern analysis and machine intelligence} \bibinfo{volume}{41} (\bibinfo{year}{2018}) \bibinfo{pages}{1979--1993}.
\bibitem[{Xie et~al.(2020)Xie, Dai, Hovy, Luong, and Le}]{xie2020unsupervised}
\bibinfo{author}{Q.~Xie}, \bibinfo{author}{Z.~Dai}, \bibinfo{author}{E.~Hovy}, \bibinfo{author}{T.~Luong}, \bibinfo{author}{Q.~Le},
\newblock \bibinfo{title}{Unsupervised data augmentation for consistency training},
\newblock \bibinfo{journal}{Advances in neural information processing systems} \bibinfo{volume}{33} (\bibinfo{year}{2020}) \bibinfo{pages}{6256--6268}.
\bibitem[{Bachman et~al.(2014)Bachman, Alsharif, and Precup}]{bachman2014learning}
\bibinfo{author}{P.~Bachman}, \bibinfo{author}{O.~Alsharif}, \bibinfo{author}{D.~Precup},
\newblock \bibinfo{title}{Learning with pseudo-ensembles},
\newblock \bibinfo{journal}{Advances in neural information processing systems} \bibinfo{volume}{27} (\bibinfo{year}{2014}).
\bibitem[{Rasmus et~al.(2015)Rasmus, Berglund, Honkala, Valpola, and Raiko}]{rasmus2015semi}
\bibinfo{author}{A.~Rasmus}, \bibinfo{author}{M.~Berglund}, \bibinfo{author}{M.~Honkala}, \bibinfo{author}{H.~Valpola}, \bibinfo{author}{T.~Raiko},
\newblock \bibinfo{title}{Semi-supervised learning with ladder networks},
\newblock \bibinfo{journal}{Advances in neural information processing systems} \bibinfo{volume}{28} (\bibinfo{year}{2015}).
\bibitem[{Berthelot et~al.(2019)Berthelot, Carlini, Goodfellow, Papernot, Oliver, and Raffel}]{berthelot2019mixmatch}
\bibinfo{author}{D.~Berthelot}, \bibinfo{author}{N.~Carlini}, \bibinfo{author}{I.~Goodfellow}, \bibinfo{author}{N.~Papernot}, \bibinfo{author}{A.~Oliver}, \bibinfo{author}{C.~A. Raffel},
\newblock \bibinfo{title}{Mixmatch: A holistic approach to semi-supervised learning},
\newblock \bibinfo{journal}{Advances in neural information processing systems} \bibinfo{volume}{32} (\bibinfo{year}{2019}).
\bibitem[{Sajjadi et~al.(2016)Sajjadi, Javanmardi, and Tasdizen}]{sajjadi2016regularization}
\bibinfo{author}{M.~Sajjadi}, \bibinfo{author}{M.~Javanmardi}, \bibinfo{author}{T.~Tasdizen},
\newblock \bibinfo{title}{Regularization with stochastic transformations and perturbations for deep semi-supervised learning},
\newblock \bibinfo{journal}{Advances in neural information processing systems} \bibinfo{volume}{29} (\bibinfo{year}{2016}).
\bibitem[{Scudder(1965)}]{scudder1965probability}
\bibinfo{author}{H.~Scudder},
\newblock \bibinfo{title}{Probability of error of some adaptive pattern-recognition machines},
\newblock \bibinfo{journal}{IEEE Transactions on Information Theory} \bibinfo{volume}{11} (\bibinfo{year}{1965}) \bibinfo{pages}{363--371}.
\bibitem[{McLachlan(1975)}]{mclachlan1975iterative}
\bibinfo{author}{G.~J. McLachlan},
\newblock \bibinfo{title}{Iterative reclassification procedure for constructing an asymptotically optimal rule of allocation in discriminant analysis},
\newblock \bibinfo{journal}{Journal of the American Statistical Association} \bibinfo{volume}{70} (\bibinfo{year}{1975}) \bibinfo{pages}{365--369}.
\bibitem[{Yarowsky(1995)}]{yarowsky1995unsupervised}
\bibinfo{author}{D.~Yarowsky},
\newblock \bibinfo{title}{Unsupervised word sense disambiguation rivaling supervised methods},
\newblock in: \bibinfo{booktitle}{33rd annual meeting of the association for computational linguistics}, \bibinfo{year}{1995}, pp. \bibinfo{pages}{189--196}.
\bibitem[{Riloff(1996)}]{riloff1996automatically}
\bibinfo{author}{E.~Riloff},
\newblock \bibinfo{title}{Automatically generating extraction patterns from untagged text},
\newblock in: \bibinfo{booktitle}{Proceedings of the national conference on artificial intelligence}, \bibinfo{year}{1996}, pp. \bibinfo{pages}{1044--1049}.
\bibitem[{Riloff and Wiebe(2003)}]{riloff2003learning}
\bibinfo{author}{E.~Riloff}, \bibinfo{author}{J.~Wiebe},
\newblock \bibinfo{title}{Learning extraction patterns for subjective expressions},
\newblock in: \bibinfo{booktitle}{Proceedings of the 2003 conference on Empirical methods in natural language processing}, \bibinfo{year}{2003}, pp. \bibinfo{pages}{105--112}.
\bibitem[{He and Zhou(2011)}]{he2011self}
\bibinfo{author}{Y.~He}, \bibinfo{author}{D.~Zhou},
\newblock \bibinfo{title}{Self-training from labeled features for sentiment analysis},
\newblock \bibinfo{journal}{Information Processing \& Management} \bibinfo{volume}{47} (\bibinfo{year}{2011}) \bibinfo{pages}{606--616}.
\bibitem[{Lee et~al.(2013)}]{lee2013pseudo}
\bibinfo{author}{D.-H. Lee}, et~al.,
\newblock \bibinfo{title}{Pseudo-label: The simple and efficient semi-supervised learning method for deep neural networks},
\newblock in: \bibinfo{booktitle}{Workshop on challenges in representation learning, ICML}, volume~\bibinfo{volume}{3}, \bibinfo{year}{2013}, p. \bibinfo{pages}{896}.
\bibitem[{Shi et~al.(2018)Shi, Gong, Ding, Tao, and Zheng}]{shi2018transductive}
\bibinfo{author}{W.~Shi}, \bibinfo{author}{Y.~Gong}, \bibinfo{author}{C.~Ding}, \bibinfo{author}{Z.~M. Tao}, \bibinfo{author}{N.~Zheng},
\newblock \bibinfo{title}{Transductive semi-supervised deep learning using min-max features},
\newblock in: \bibinfo{booktitle}{Proceedings of the European Conference on Computer Vision (ECCV)}, \bibinfo{year}{2018}, pp. \bibinfo{pages}{299--315}.
\bibitem[{Iscen et~al.(2019)Iscen, Tolias, Avrithis, and Chum}]{iscen2019label}
\bibinfo{author}{A.~Iscen}, \bibinfo{author}{G.~Tolias}, \bibinfo{author}{Y.~Avrithis}, \bibinfo{author}{O.~Chum},
\newblock \bibinfo{title}{Label propagation for deep semi-supervised learning},
\newblock in: \bibinfo{booktitle}{Proceedings of the IEEE/CVF conference on computer vision and pattern recognition}, \bibinfo{year}{2019}, pp. \bibinfo{pages}{5070--5079}.
\bibitem[{Arazo et~al.(2020)Arazo, Ortego, Albert, O’Connor, and McGuinness}]{arazo2020pseudo}
\bibinfo{author}{E.~Arazo}, \bibinfo{author}{D.~Ortego}, \bibinfo{author}{P.~Albert}, \bibinfo{author}{N.~E. O’Connor}, \bibinfo{author}{K.~McGuinness},
\newblock \bibinfo{title}{Pseudo-labeling and confirmation bias in deep semi-supervised learning},
\newblock in: \bibinfo{booktitle}{2020 International joint conference on neural networks (IJCNN)}, \bibinfo{organization}{IEEE}, \bibinfo{year}{2020}, pp. \bibinfo{pages}{1--8}.
\bibitem[{Sohn et~al.(2020)Sohn, Zhang, Li, Zhang, Lee, and Pfister}]{sohn2020simple}
\bibinfo{author}{K.~Sohn}, \bibinfo{author}{Z.~Zhang}, \bibinfo{author}{C.-L. Li}, \bibinfo{author}{H.~Zhang}, \bibinfo{author}{C.-Y. Lee}, \bibinfo{author}{T.~Pfister},
\newblock \bibinfo{title}{A simple semi-supervised learning framework for object detection},
\newblock \bibinfo{journal}{arXiv preprint arXiv:2005.04757}  (\bibinfo{year}{2020}).
\bibitem[{Xie et~al.(2020)Xie, Luong, Hovy, and Le}]{xie2020self}
\bibinfo{author}{Q.~Xie}, \bibinfo{author}{M.-T. Luong}, \bibinfo{author}{E.~Hovy}, \bibinfo{author}{Q.~V. Le},
\newblock \bibinfo{title}{Self-training with noisy student improves imagenet classification},
\newblock in: \bibinfo{booktitle}{Proceedings of the IEEE/CVF conference on computer vision and pattern recognition}, \bibinfo{year}{2020}, pp. \bibinfo{pages}{10687--10698}.
\bibitem[{Pham et~al.(2021)Pham, Dai, Xie, and Le}]{pham2021meta}
\bibinfo{author}{H.~Pham}, \bibinfo{author}{Z.~Dai}, \bibinfo{author}{Q.~Xie}, \bibinfo{author}{Q.~V. Le},
\newblock \bibinfo{title}{Meta pseudo labels},
\newblock in: \bibinfo{booktitle}{Proceedings of the IEEE/CVF conference on computer vision and pattern recognition}, \bibinfo{year}{2021}, pp. \bibinfo{pages}{11557--11568}.
\bibitem[{Cai et~al.(2021)Cai, Ravichandran, Maji, Fowlkes, Tu, and Soatto}]{cai2021exponential}
\bibinfo{author}{Z.~Cai}, \bibinfo{author}{A.~Ravichandran}, \bibinfo{author}{S.~Maji}, \bibinfo{author}{C.~Fowlkes}, \bibinfo{author}{Z.~Tu}, \bibinfo{author}{S.~Soatto},
\newblock \bibinfo{title}{Exponential moving average normalization for self-supervised and semi-supervised learning},
\newblock in: \bibinfo{booktitle}{Proceedings of the IEEE/CVF Conference on Computer Vision and Pattern Recognition}, \bibinfo{year}{2021}, pp. \bibinfo{pages}{194--203}.
\bibitem[{Liu et~al.(2021)Liu, Ma, He, Kuo, Chen, Zhang, Wu, Kira, and Vajda}]{liu2021unbiased}
\bibinfo{author}{Y.-C. Liu}, \bibinfo{author}{C.-Y. Ma}, \bibinfo{author}{Z.~He}, \bibinfo{author}{C.-W. Kuo}, \bibinfo{author}{K.~Chen}, \bibinfo{author}{P.~Zhang}, \bibinfo{author}{B.~Wu}, \bibinfo{author}{Z.~Kira}, \bibinfo{author}{P.~Vajda},
\newblock \bibinfo{title}{Unbiased teacher for semi-supervised object detection},
\newblock \bibinfo{journal}{arXiv preprint arXiv:2102.09480}  (\bibinfo{year}{2021}).
\bibitem[{Zhang et~al.(2021)Zhang, Wang, Hou, Wu, Wang, Okumura, and Shinozaki}]{zhang2021flexmatch}
\bibinfo{author}{B.~Zhang}, \bibinfo{author}{Y.~Wang}, \bibinfo{author}{W.~Hou}, \bibinfo{author}{H.~Wu}, \bibinfo{author}{J.~Wang}, \bibinfo{author}{M.~Okumura}, \bibinfo{author}{T.~Shinozaki},
\newblock \bibinfo{title}{Flexmatch: Boosting semi-supervised learning with curriculum pseudo labeling},
\newblock \bibinfo{journal}{Advances in Neural Information Processing Systems} \bibinfo{volume}{34} (\bibinfo{year}{2021}) \bibinfo{pages}{18408--18419}.
\bibitem[{Wei et~al.(2021)Wei, Sohn, Mellina, Yuille, and Yang}]{wei2021crest}
\bibinfo{author}{C.~Wei}, \bibinfo{author}{K.~Sohn}, \bibinfo{author}{C.~Mellina}, \bibinfo{author}{A.~Yuille}, \bibinfo{author}{F.~Yang},
\newblock \bibinfo{title}{Crest: A class-rebalancing self-training framework for imbalanced semi-supervised learning},
\newblock in: \bibinfo{booktitle}{Proceedings of the IEEE/CVF conference on computer vision and pattern recognition}, \bibinfo{year}{2021}, pp. \bibinfo{pages}{10857--10866}.
\bibitem[{Pedronette and Latecki(2021)}]{pedronette2021rank}
\bibinfo{author}{D.~C.~G. Pedronette}, \bibinfo{author}{L.~J. Latecki},
\newblock \bibinfo{title}{Rank-based self-training for graph convolutional networks},
\newblock \bibinfo{journal}{Information Processing \& Management} \bibinfo{volume}{58} (\bibinfo{year}{2021}) \bibinfo{pages}{102443}.
\bibitem[{Alqahtani et~al.(2023)Alqahtani, Al-Twairesh, and Alsanad}]{alqahtani2023improving}
\bibinfo{author}{Y.~Alqahtani}, \bibinfo{author}{N.~Al-Twairesh}, \bibinfo{author}{A.~Alsanad},
\newblock \bibinfo{title}{Improving sentiment domain adaptation for arabic using an unsupervised self-labeling framework},
\newblock \bibinfo{journal}{Information Processing \& Management} \bibinfo{volume}{60} (\bibinfo{year}{2023}) \bibinfo{pages}{103338}.
\bibitem[{Cai et~al.(2022)Cai, Ravichandran, Favaro, Wang, Modolo, Bhotika, Tu, and Soatto}]{cai2022semi}
\bibinfo{author}{Z.~Cai}, \bibinfo{author}{A.~Ravichandran}, \bibinfo{author}{P.~Favaro}, \bibinfo{author}{M.~Wang}, \bibinfo{author}{D.~Modolo}, \bibinfo{author}{R.~Bhotika}, \bibinfo{author}{Z.~Tu}, \bibinfo{author}{S.~Soatto},
\newblock \bibinfo{title}{Semi-supervised vision transformers at scale},
\newblock \bibinfo{journal}{Advances in Neural Information Processing Systems} \bibinfo{volume}{35} (\bibinfo{year}{2022}) \bibinfo{pages}{25697--25710}.
\bibitem[{Wang et~al.(2023)Wang, Chen, Heng, Hou, Fan, , Wu, Wang, Savvides, Shinozaki, Raj, Schiele, and Xie}]{wang2023freematch}
\bibinfo{author}{Y.~Wang}, \bibinfo{author}{H.~Chen}, \bibinfo{author}{Q.~Heng}, \bibinfo{author}{W.~Hou}, \bibinfo{author}{Y.~Fan}, , \bibinfo{author}{Z.~Wu}, \bibinfo{author}{J.~Wang}, \bibinfo{author}{M.~Savvides}, \bibinfo{author}{T.~Shinozaki}, \bibinfo{author}{B.~Raj}, \bibinfo{author}{B.~Schiele}, \bibinfo{author}{X.~Xie},
\newblock \bibinfo{title}{Freematch: Self-adaptive thresholding for semi-supervised learning},
\newblock \bibinfo{journal}{International Conference on Learning Representations (ICLR)}  (\bibinfo{year}{2023}).
\bibitem[{Chen et~al.(2022)Chen, Jiang, Wang, Wan, Wang, and Long}]{chen2022debiased}
\bibinfo{author}{B.~Chen}, \bibinfo{author}{J.~Jiang}, \bibinfo{author}{X.~Wang}, \bibinfo{author}{P.~Wan}, \bibinfo{author}{J.~Wang}, \bibinfo{author}{M.~Long},
\newblock \bibinfo{title}{Debiased self-training for semi-supervised learning},
\newblock \bibinfo{journal}{Advances in Neural Information Processing Systems} \bibinfo{volume}{35} (\bibinfo{year}{2022}) \bibinfo{pages}{32424--32437}.
\bibitem[{Wang et~al.(2022)Wang, Wu, Lian, and Yu}]{wang2022debiased}
\bibinfo{author}{X.~Wang}, \bibinfo{author}{Z.~Wu}, \bibinfo{author}{L.~Lian}, \bibinfo{author}{S.~X. Yu},
\newblock \bibinfo{title}{Debiased learning from naturally imbalanced pseudo-labels},
\newblock in: \bibinfo{booktitle}{Proceedings of the IEEE/CVF Conference on Computer Vision and Pattern Recognition}, \bibinfo{year}{2022}, pp. \bibinfo{pages}{14647--14657}.
\bibitem[{Bossard et~al.(2014)Bossard, Guillaumin, and Van~Gool}]{bossard2014food}
\bibinfo{author}{L.~Bossard}, \bibinfo{author}{M.~Guillaumin}, \bibinfo{author}{L.~Van~Gool},
\newblock \bibinfo{title}{Food-101--mining discriminative components with random forests},
\newblock in: \bibinfo{booktitle}{Computer Vision--ECCV 2014: 13th European Conference, Zurich, Switzerland, September 6-12, 2014, Proceedings, Part VI 13}, \bibinfo{organization}{Springer}, \bibinfo{year}{2014}, pp. \bibinfo{pages}{446--461}.
\bibitem[{Grant Van~Horn(2019)}]{inaturalist-2019-fgvc6}
\bibinfo{author}{M.~W.~K. Grant Van~Horn, macaodha}, \bibinfo{title}{inaturalist 2019 at fgvc6}, \bibinfo{year}{2019}. \URLprefix \url{https://kaggle.com/competitions/inaturalist-2019-fgvc6}.
\bibitem[{Krizhevsky et~al.(2009)Krizhevsky, Hinton et~al.}]{krizhevsky2009learning}
\bibinfo{author}{A.~Krizhevsky}, \bibinfo{author}{G.~Hinton}, et~al.,
\newblock \bibinfo{title}{Learning multiple layers of features from tiny images}  (\bibinfo{year}{2009}).
\bibitem[{Netzer et~al.(2011)Netzer, Wang, Coates, Bissacco, Wu, Ng et~al.}]{netzer2011reading}
\bibinfo{author}{Y.~Netzer}, \bibinfo{author}{T.~Wang}, \bibinfo{author}{A.~Coates}, \bibinfo{author}{A.~Bissacco}, \bibinfo{author}{B.~Wu}, \bibinfo{author}{A.~Y. Ng}, et~al.,
\newblock \bibinfo{title}{Reading digits in natural images with unsupervised feature learning},
\newblock in: \bibinfo{booktitle}{NIPS workshop on deep learning and unsupervised feature learning}, volume \bibinfo{volume}{2011}, \bibinfo{organization}{Granada}, \bibinfo{year}{2011}, p.~\bibinfo{pages}{4}.
\bibitem[{Coates et~al.(2011)Coates, Ng, and Lee}]{coates2011analysis}
\bibinfo{author}{A.~Coates}, \bibinfo{author}{A.~Ng}, \bibinfo{author}{H.~Lee},
\newblock \bibinfo{title}{An analysis of single-layer networks in unsupervised feature learning},
\newblock in: \bibinfo{booktitle}{Proceedings of the fourteenth international conference on artificial intelligence and statistics}, \bibinfo{organization}{JMLR Workshop and Conference Proceedings}, \bibinfo{year}{2011}, pp. \bibinfo{pages}{215--223}.
\bibitem[{Xiao et~al.(2015)Xiao, Xia, Yang, Huang, and Wang}]{xiao2015learning}
\bibinfo{author}{T.~Xiao}, \bibinfo{author}{T.~Xia}, \bibinfo{author}{Y.~Yang}, \bibinfo{author}{C.~Huang}, \bibinfo{author}{X.~Wang},
\newblock \bibinfo{title}{Learning from massive noisy labeled data for image classification},
\newblock in: \bibinfo{booktitle}{Proceedings of the IEEE conference on computer vision and pattern recognition}, \bibinfo{year}{2015}, pp. \bibinfo{pages}{2691--2699}.
\bibitem[{Dosovitskiy et~al.(2020)Dosovitskiy, Beyer, Kolesnikov, Weissenborn, Zhai, Unterthiner, Dehghani, Minderer, Heigold, Gelly et~al.}]{dosovitskiy2020image}
\bibinfo{author}{A.~Dosovitskiy}, \bibinfo{author}{L.~Beyer}, \bibinfo{author}{A.~Kolesnikov}, \bibinfo{author}{D.~Weissenborn}, \bibinfo{author}{X.~Zhai}, \bibinfo{author}{T.~Unterthiner}, \bibinfo{author}{M.~Dehghani}, \bibinfo{author}{M.~Minderer}, \bibinfo{author}{G.~Heigold}, \bibinfo{author}{S.~Gelly}, et~al.,
\newblock \bibinfo{title}{An image is worth 16x16 words: Transformers for image recognition at scale},
\newblock \bibinfo{journal}{arXiv preprint arXiv:2010.11929}  (\bibinfo{year}{2020}).
\bibitem[{Caron et~al.(2021)Caron, Touvron, Misra, J{\'e}gou, Mairal, Bojanowski, and Joulin}]{caron2021emerging}
\bibinfo{author}{M.~Caron}, \bibinfo{author}{H.~Touvron}, \bibinfo{author}{I.~Misra}, \bibinfo{author}{H.~J{\'e}gou}, \bibinfo{author}{J.~Mairal}, \bibinfo{author}{P.~Bojanowski}, \bibinfo{author}{A.~Joulin},
\newblock \bibinfo{title}{Emerging properties in self-supervised vision transformers},
\newblock in: \bibinfo{booktitle}{Proceedings of the IEEE/CVF international conference on computer vision}, \bibinfo{year}{2021}, pp. \bibinfo{pages}{9650--9660}.
\bibitem[{He et~al.(2022)He, Chen, Xie, Li, Doll{\'a}r, and Girshick}]{he2022masked}
\bibinfo{author}{K.~He}, \bibinfo{author}{X.~Chen}, \bibinfo{author}{S.~Xie}, \bibinfo{author}{Y.~Li}, \bibinfo{author}{P.~Doll{\'a}r}, \bibinfo{author}{R.~Girshick},
\newblock \bibinfo{title}{Masked autoencoders are scalable vision learners},
\newblock in: \bibinfo{booktitle}{Proceedings of the IEEE/CVF conference on computer vision and pattern recognition}, \bibinfo{year}{2022}, pp. \bibinfo{pages}{16000--16009}.
\bibitem[{Liu et~al.(2022)Liu, Mao, Wu, Feichtenhofer, Darrell, and Xie}]{liu2022convnet}
\bibinfo{author}{Z.~Liu}, \bibinfo{author}{H.~Mao}, \bibinfo{author}{C.-Y. Wu}, \bibinfo{author}{C.~Feichtenhofer}, \bibinfo{author}{T.~Darrell}, \bibinfo{author}{S.~Xie},
\newblock \bibinfo{title}{A convnet for the 2020s},
\newblock in: \bibinfo{booktitle}{Proceedings of the IEEE/CVF conference on computer vision and pattern recognition}, \bibinfo{year}{2022}, pp. \bibinfo{pages}{11976--11986}.
\bibitem[{Zhang et~al.(2017)Zhang, Cisse, Dauphin, and Lopez-Paz}]{zhang2017mixup}
\bibinfo{author}{H.~Zhang}, \bibinfo{author}{M.~Cisse}, \bibinfo{author}{Y.~N. Dauphin}, \bibinfo{author}{D.~Lopez-Paz},
\newblock \bibinfo{title}{mixup: Beyond empirical risk minimization},
\newblock \bibinfo{journal}{arXiv preprint arXiv:1710.09412}  (\bibinfo{year}{2017}).
\bibitem[{He et~al.(2016)He, Zhang, Ren, and Sun}]{he2016deep}
\bibinfo{author}{K.~He}, \bibinfo{author}{X.~Zhang}, \bibinfo{author}{S.~Ren}, \bibinfo{author}{J.~Sun},
\newblock \bibinfo{title}{Deep residual learning for image recognition},
\newblock in: \bibinfo{booktitle}{Proceedings of the IEEE conference on computer vision and pattern recognition}, \bibinfo{year}{2016}, pp. \bibinfo{pages}{770--778}.
\bibitem[{Loshchilov and Hutter(2017)}]{loshchilov2017decoupled}
\bibinfo{author}{I.~Loshchilov}, \bibinfo{author}{F.~Hutter},
\newblock \bibinfo{title}{Decoupled weight decay regularization},
\newblock \bibinfo{journal}{arXiv preprint arXiv:1711.05101}  (\bibinfo{year}{2017}).
\bibitem[{Goyal et~al.(2017)Goyal, Doll{\'a}r, Girshick, Noordhuis, Wesolowski, Kyrola, Tulloch, Jia, and He}]{goyal2017accurate}
\bibinfo{author}{P.~Goyal}, \bibinfo{author}{P.~Doll{\'a}r}, \bibinfo{author}{R.~Girshick}, \bibinfo{author}{P.~Noordhuis}, \bibinfo{author}{L.~Wesolowski}, \bibinfo{author}{A.~Kyrola}, \bibinfo{author}{A.~Tulloch}, \bibinfo{author}{Y.~Jia}, \bibinfo{author}{K.~He},
\newblock \bibinfo{title}{Accurate, large minibatch sgd: Training imagenet in 1 hour},
\newblock \bibinfo{journal}{arXiv preprint arXiv:1706.02677}  (\bibinfo{year}{2017}).
\bibitem[{Loshchilov and Hutter(2016)}]{loshchilov2016sgdr}
\bibinfo{author}{I.~Loshchilov}, \bibinfo{author}{F.~Hutter},
\newblock \bibinfo{title}{Sgdr: Stochastic gradient descent with warm restarts},
\newblock \bibinfo{journal}{arXiv preprint arXiv:1608.03983}  (\bibinfo{year}{2016}).
\bibitem[{Szegedy et~al.(2016)Szegedy, Vanhoucke, Ioffe, Shlens, and Wojna}]{szegedy2016rethinking}
\bibinfo{author}{C.~Szegedy}, \bibinfo{author}{V.~Vanhoucke}, \bibinfo{author}{S.~Ioffe}, \bibinfo{author}{J.~Shlens}, \bibinfo{author}{Z.~Wojna},
\newblock \bibinfo{title}{Rethinking the inception architecture for computer vision},
\newblock in: \bibinfo{booktitle}{Proceedings of the IEEE conference on computer vision and pattern recognition}, \bibinfo{year}{2016}, pp. \bibinfo{pages}{2818--2826}.
\bibitem[{Huang et~al.(2016)Huang, Sun, Liu, Sedra, and Weinberger}]{huang2016deep}
\bibinfo{author}{G.~Huang}, \bibinfo{author}{Y.~Sun}, \bibinfo{author}{Z.~Liu}, \bibinfo{author}{D.~Sedra}, \bibinfo{author}{K.~Q. Weinberger},
\newblock \bibinfo{title}{Deep networks with stochastic depth},
\newblock in: \bibinfo{booktitle}{Computer Vision--ECCV 2016: 14th European Conference, Amsterdam, The Netherlands, October 11--14, 2016, Proceedings, Part IV 14}, \bibinfo{organization}{Springer}, \bibinfo{year}{2016}, pp. \bibinfo{pages}{646--661}.
\bibitem[{Yun et~al.(2019)Yun, Han, Oh, Chun, Choe, and Yoo}]{yun2019cutmix}
\bibinfo{author}{S.~Yun}, \bibinfo{author}{D.~Han}, \bibinfo{author}{S.~J. Oh}, \bibinfo{author}{S.~Chun}, \bibinfo{author}{J.~Choe}, \bibinfo{author}{Y.~Yoo},
\newblock \bibinfo{title}{Cutmix: Regularization strategy to train strong classifiers with localizable features},
\newblock in: \bibinfo{booktitle}{Proceedings of the IEEE/CVF international conference on computer vision}, \bibinfo{year}{2019}, pp. \bibinfo{pages}{6023--6032}.
\bibitem[{Cubuk et~al.(2020)Cubuk, Zoph, Shlens, and Le}]{cubuk2020randaugment}
\bibinfo{author}{E.~D. Cubuk}, \bibinfo{author}{B.~Zoph}, \bibinfo{author}{J.~Shlens}, \bibinfo{author}{Q.~V. Le},
\newblock \bibinfo{title}{Randaugment: Practical automated data augmentation with a reduced search space},
\newblock in: \bibinfo{booktitle}{Proceedings of the IEEE/CVF conference on computer vision and pattern recognition workshops}, \bibinfo{year}{2020}, pp. \bibinfo{pages}{702--703}.
\bibitem[{Zhong et~al.(2020)Zhong, Zheng, Kang, Li, and Yang}]{zhong2020random}
\bibinfo{author}{Z.~Zhong}, \bibinfo{author}{L.~Zheng}, \bibinfo{author}{G.~Kang}, \bibinfo{author}{S.~Li}, \bibinfo{author}{Y.~Yang},
\newblock \bibinfo{title}{Random erasing data augmentation},
\newblock in: \bibinfo{booktitle}{Proceedings of the AAAI conference on artificial intelligence}, volume~\bibinfo{volume}{34}, \bibinfo{year}{2020}, pp. \bibinfo{pages}{13001--13008}.
\bibitem[{Chen et~al.(2020)Chen, Kornblith, Swersky, Norouzi, and Hinton}]{chen2020big}
\bibinfo{author}{T.~Chen}, \bibinfo{author}{S.~Kornblith}, \bibinfo{author}{K.~Swersky}, \bibinfo{author}{M.~Norouzi}, \bibinfo{author}{G.~E. Hinton},
\newblock \bibinfo{title}{Big self-supervised models are strong semi-supervised learners},
\newblock \bibinfo{journal}{Advances in neural information processing systems} \bibinfo{volume}{33} (\bibinfo{year}{2020}) \bibinfo{pages}{22243--22255}.
\bibitem[{Li et~al.(2021)Li, Xiong, and Hoi}]{li2021comatch}
\bibinfo{author}{J.~Li}, \bibinfo{author}{C.~Xiong}, \bibinfo{author}{S.~C. Hoi},
\newblock \bibinfo{title}{Comatch: Semi-supervised learning with contrastive graph regularization},
\newblock in: \bibinfo{booktitle}{Proceedings of the IEEE/CVF international conference on computer vision}, \bibinfo{year}{2021}, pp. \bibinfo{pages}{9475--9484}.
\bibitem[{Assran et~al.(2021)Assran, Caron, Misra, Bojanowski, Joulin, Ballas, and Rabbat}]{assran2021semi}
\bibinfo{author}{M.~Assran}, \bibinfo{author}{M.~Caron}, \bibinfo{author}{I.~Misra}, \bibinfo{author}{P.~Bojanowski}, \bibinfo{author}{A.~Joulin}, \bibinfo{author}{N.~Ballas}, \bibinfo{author}{M.~Rabbat},
\newblock \bibinfo{title}{Semi-supervised learning of visual features by non-parametrically predicting view assignments with support samples},
\newblock in: \bibinfo{booktitle}{Proceedings of the IEEE/CVF International Conference on Computer Vision}, \bibinfo{year}{2021}, pp. \bibinfo{pages}{8443--8452}.
\bibitem[{Zheng et~al.(2022)Zheng, You, Huang, Wang, Qian, and Xu}]{zheng2022simmatch}
\bibinfo{author}{M.~Zheng}, \bibinfo{author}{S.~You}, \bibinfo{author}{L.~Huang}, \bibinfo{author}{F.~Wang}, \bibinfo{author}{C.~Qian}, \bibinfo{author}{C.~Xu},
\newblock \bibinfo{title}{Simmatch: Semi-supervised learning with similarity matching},
\newblock in: \bibinfo{booktitle}{Proceedings of the IEEE/CVF Conference on Computer Vision and Pattern Recognition}, \bibinfo{year}{2022}, pp. \bibinfo{pages}{14471--14481}.
\bibitem[{Zheng et~al.(2023)Zheng, You, Huang, Luo, Wang, Qian, and Xu}]{zheng2023simmatchv2}
\bibinfo{author}{M.~Zheng}, \bibinfo{author}{S.~You}, \bibinfo{author}{L.~Huang}, \bibinfo{author}{C.~Luo}, \bibinfo{author}{F.~Wang}, \bibinfo{author}{C.~Qian}, \bibinfo{author}{C.~Xu},
\newblock \bibinfo{title}{Simmatchv2: Semi-supervised learning with graph consistency},
\newblock in: \bibinfo{booktitle}{Proceedings of the IEEE/CVF International Conference on Computer Vision}, \bibinfo{year}{2023}, pp. \bibinfo{pages}{16432--16442}.
\bibitem[{Weng et~al.(2022)Weng, Yang, Li, Wu, and Jiang}]{weng2022semi}
\bibinfo{author}{Z.~Weng}, \bibinfo{author}{X.~Yang}, \bibinfo{author}{A.~Li}, \bibinfo{author}{Z.~Wu}, \bibinfo{author}{Y.-G. Jiang},
\newblock \bibinfo{title}{Semi-supervised vision transformers},
\newblock in: \bibinfo{booktitle}{European conference on computer vision}, \bibinfo{organization}{Springer Nature Switzerland Cham}, \bibinfo{year}{2022}, pp. \bibinfo{pages}{605--620}.
\bibitem[{Hinton et~al.(2015)Hinton, Vinyals, and Dean}]{hinton2015distilling}
\bibinfo{author}{G.~Hinton}, \bibinfo{author}{O.~Vinyals}, \bibinfo{author}{J.~Dean},
\newblock \bibinfo{title}{Distilling the knowledge in a neural network},
\newblock \bibinfo{journal}{arXiv preprint arXiv:1503.02531}  (\bibinfo{year}{2015}).
\bibitem[{Mirzadeh et~al.(2020)Mirzadeh, Farajtabar, Li, Levine, Matsukawa, and Ghasemzadeh}]{mirzadeh2020improved}
\bibinfo{author}{S.~I. Mirzadeh}, \bibinfo{author}{M.~Farajtabar}, \bibinfo{author}{A.~Li}, \bibinfo{author}{N.~Levine}, \bibinfo{author}{A.~Matsukawa}, \bibinfo{author}{H.~Ghasemzadeh},
\newblock \bibinfo{title}{Improved knowledge distillation via teacher assistant},
\newblock in: \bibinfo{booktitle}{Proceedings of the AAAI conference on artificial intelligence}, volume~\bibinfo{volume}{34}, \bibinfo{year}{2020}, pp. \bibinfo{pages}{5191--5198}.
\bibitem[{Chen et~al.(2021)Chen, Mei, Zhang, Wang, Wang, Feng, and Chen}]{chen2021cross}
\bibinfo{author}{D.~Chen}, \bibinfo{author}{J.-P. Mei}, \bibinfo{author}{Y.~Zhang}, \bibinfo{author}{C.~Wang}, \bibinfo{author}{Z.~Wang}, \bibinfo{author}{Y.~Feng}, \bibinfo{author}{C.~Chen},
\newblock \bibinfo{title}{Cross-layer distillation with semantic calibration},
\newblock in: \bibinfo{booktitle}{Proceedings of the AAAI conference on artificial intelligence}, volume~\bibinfo{volume}{35}, \bibinfo{year}{2021}, pp. \bibinfo{pages}{7028--7036}.
\bibitem[{Passalis and Tefas(2018)}]{passalis2018learning}
\bibinfo{author}{N.~Passalis}, \bibinfo{author}{A.~Tefas},
\newblock \bibinfo{title}{Learning deep representations with probabilistic knowledge transfer},
\newblock in: \bibinfo{booktitle}{Proceedings of the European Conference on Computer Vision (ECCV)}, \bibinfo{year}{2018}, pp. \bibinfo{pages}{268--284}.
\bibitem[{Park et~al.(2019)Park, Kim, Lu, and Cho}]{park2019relational}
\bibinfo{author}{W.~Park}, \bibinfo{author}{D.~Kim}, \bibinfo{author}{Y.~Lu}, \bibinfo{author}{M.~Cho},
\newblock \bibinfo{title}{Relational knowledge distillation},
\newblock in: \bibinfo{booktitle}{Proceedings of the IEEE/CVF conference on computer vision and pattern recognition}, \bibinfo{year}{2019}, pp. \bibinfo{pages}{3967--3976}.
\bibitem[{Zhu et~al.(2021)Zhu, Tang, Chen, Yu, Liu, Rong, Yang, and Wang}]{zhu2021complementary}
\bibinfo{author}{J.~Zhu}, \bibinfo{author}{S.~Tang}, \bibinfo{author}{D.~Chen}, \bibinfo{author}{S.~Yu}, \bibinfo{author}{Y.~Liu}, \bibinfo{author}{M.~Rong}, \bibinfo{author}{A.~Yang}, \bibinfo{author}{X.~Wang},
\newblock \bibinfo{title}{Complementary relation contrastive distillation},
\newblock in: \bibinfo{booktitle}{Proceedings of the IEEE/CVF conference on computer vision and pattern recognition}, \bibinfo{year}{2021}, pp. \bibinfo{pages}{9260--9269}.
\bibitem[{Yang et~al.(2022)Yang, An, Zhou, Cai, Zhi, Wu, Xu, and Zhang}]{yang2022mixskd}
\bibinfo{author}{C.~Yang}, \bibinfo{author}{Z.~An}, \bibinfo{author}{H.~Zhou}, \bibinfo{author}{L.~Cai}, \bibinfo{author}{X.~Zhi}, \bibinfo{author}{J.~Wu}, \bibinfo{author}{Y.~Xu}, \bibinfo{author}{Q.~Zhang},
\newblock \bibinfo{title}{Mixskd: Self-knowledge distillation from mixup for image recognition},
\newblock in: \bibinfo{booktitle}{European Conference on Computer Vision}, \bibinfo{organization}{Springer}, \bibinfo{year}{2022}, pp. \bibinfo{pages}{534--551}.
\bibitem[{Wang et~al.(2022)Wang, Chen, Fan, Sun, Tao, Hou, Wang, Yang, Zhou, Guo et~al.}]{wang2022usb}
\bibinfo{author}{Y.~Wang}, \bibinfo{author}{H.~Chen}, \bibinfo{author}{Y.~Fan}, \bibinfo{author}{W.~Sun}, \bibinfo{author}{R.~Tao}, \bibinfo{author}{W.~Hou}, \bibinfo{author}{R.~Wang}, \bibinfo{author}{L.~Yang}, \bibinfo{author}{Z.~Zhou}, \bibinfo{author}{L.-Z. Guo}, et~al.,
\newblock \bibinfo{title}{Usb: A unified semi-supervised learning benchmark for classification},
\newblock \bibinfo{journal}{Advances in Neural Information Processing Systems} \bibinfo{volume}{35} (\bibinfo{year}{2022}) \bibinfo{pages}{3938--3961}.
\bibitem[{Touvron et~al.(2021)Touvron, Cord, Douze, Massa, Sablayrolles, and J{\'e}gou}]{touvron2021training}
\bibinfo{author}{H.~Touvron}, \bibinfo{author}{M.~Cord}, \bibinfo{author}{M.~Douze}, \bibinfo{author}{F.~Massa}, \bibinfo{author}{A.~Sablayrolles}, \bibinfo{author}{H.~J{\'e}gou},
\newblock \bibinfo{title}{Training data-efficient image transformers \& distillation through attention},
\newblock in: \bibinfo{booktitle}{International conference on machine learning}, \bibinfo{organization}{PMLR}, \bibinfo{year}{2021}, pp. \bibinfo{pages}{10347--10357}.
\bibitem[{Touvron et~al.(2022)Touvron, Cord, and J{\'e}gou}]{touvron2022deit}
\bibinfo{author}{H.~Touvron}, \bibinfo{author}{M.~Cord}, \bibinfo{author}{H.~J{\'e}gou},
\newblock \bibinfo{title}{Deit iii: Revenge of the vit},
\newblock in: \bibinfo{booktitle}{European conference on computer vision}, \bibinfo{organization}{Springer Nature Switzerland Cham}, \bibinfo{year}{2022}, pp. \bibinfo{pages}{516--533}.
\bibitem[{Chen et~al.(2023)Chen, Tao, Fan, Wang, Wang, Schiele, Xie, Raj, and Savvides}]{chen2023softmatch}
\bibinfo{author}{H.~Chen}, \bibinfo{author}{R.~Tao}, \bibinfo{author}{Y.~Fan}, \bibinfo{author}{Y.~Wang}, \bibinfo{author}{J.~Wang}, \bibinfo{author}{B.~Schiele}, \bibinfo{author}{X.~Xie}, \bibinfo{author}{B.~Raj}, \bibinfo{author}{M.~Savvides},
\newblock \bibinfo{title}{Softmatch: Addressing the quantity-quality trade-off in semi-supervised learning},
\newblock \bibinfo{journal}{International Conference on Learning Representations (ICLR)}  (\bibinfo{year}{2023}).
\bibitem[{Zagoruyko and Komodakis(2016)}]{zagoruyko2016wide}
\bibinfo{author}{S.~Zagoruyko}, \bibinfo{author}{N.~Komodakis},
\newblock \bibinfo{title}{Wide residual networks},
\newblock \bibinfo{journal}{arXiv preprint arXiv:1605.07146}  (\bibinfo{year}{2016}).
\bibitem[{Xu et~al.(2021)Xu, Shang, Ye, Qian, Li, Sun, Li, and Jin}]{xu2021dash}
\bibinfo{author}{Y.~Xu}, \bibinfo{author}{L.~Shang}, \bibinfo{author}{J.~Ye}, \bibinfo{author}{Q.~Qian}, \bibinfo{author}{Y.-F. Li}, \bibinfo{author}{B.~Sun}, \bibinfo{author}{H.~Li}, \bibinfo{author}{R.~Jin},
\newblock \bibinfo{title}{Dash: Semi-supervised learning with dynamic thresholding},
\newblock in: \bibinfo{booktitle}{International Conference on Machine Learning}, \bibinfo{organization}{PMLR}, \bibinfo{year}{2021}, pp. \bibinfo{pages}{11525--11536}.
\bibitem[{Berthelot et~al.(2021)Berthelot, Roelofs, Sohn, Carlini, and Kurakin}]{berthelot2021adamatch}
\bibinfo{author}{D.~Berthelot}, \bibinfo{author}{R.~Roelofs}, \bibinfo{author}{K.~Sohn}, \bibinfo{author}{N.~Carlini}, \bibinfo{author}{A.~Kurakin},
\newblock \bibinfo{title}{Adamatch: A unified approach to semi-supervised learning and domain adaptation},
\newblock \bibinfo{journal}{arXiv preprint arXiv:2106.04732}  (\bibinfo{year}{2021}).
\bibitem[{Guo and Li(2022)}]{guo2022class}
\bibinfo{author}{L.-Z. Guo}, \bibinfo{author}{Y.-F. Li},
\newblock \bibinfo{title}{Class-imbalanced semi-supervised learning with adaptive thresholding},
\newblock in: \bibinfo{booktitle}{International Conference on Machine Learning}, \bibinfo{organization}{PMLR}, \bibinfo{year}{2022}, pp. \bibinfo{pages}{8082--8094}.
\bibitem[{Lin et~al.(2019)Lin, Shou, and Chang}]{lin2019towards}
\bibinfo{author}{X.~Lin}, \bibinfo{author}{Z.~Shou}, \bibinfo{author}{S.-F. Chang},
\newblock \bibinfo{title}{Towards train-test consistency for semi-supervised temporal action localization},
\newblock \bibinfo{journal}{arXiv preprint arXiv:1910.11285}  (\bibinfo{year}{2019}).
\bibitem[{Chen et~al.(2020)Chen, Wei, Kumar, and Ma}]{chen2020self}
\bibinfo{author}{Y.~Chen}, \bibinfo{author}{C.~Wei}, \bibinfo{author}{A.~Kumar}, \bibinfo{author}{T.~Ma},
\newblock \bibinfo{title}{Self-training avoids using spurious features under domain shift},
\newblock \bibinfo{journal}{Advances in Neural Information Processing Systems} \bibinfo{volume}{33} (\bibinfo{year}{2020}) \bibinfo{pages}{21061--21071}.
\bibitem[{Kumar et~al.(2020)Kumar, Ma, and Liang}]{kumar2020understanding}
\bibinfo{author}{A.~Kumar}, \bibinfo{author}{T.~Ma}, \bibinfo{author}{P.~Liang},
\newblock \bibinfo{title}{Understanding self-training for gradual domain adaptation},
\newblock in: \bibinfo{booktitle}{International conference on machine learning}, \bibinfo{organization}{PMLR}, \bibinfo{year}{2020}, pp. \bibinfo{pages}{5468--5479}.
\bibitem[{Zou et~al.(2018)Zou, Yu, Kumar, and Wang}]{zou2018unsupervised}
\bibinfo{author}{Y.~Zou}, \bibinfo{author}{Z.~Yu}, \bibinfo{author}{B.~Kumar}, \bibinfo{author}{J.~Wang},
\newblock \bibinfo{title}{Unsupervised domain adaptation for semantic segmentation via class-balanced self-training},
\newblock in: \bibinfo{booktitle}{Proceedings of the European conference on computer vision (ECCV)}, \bibinfo{year}{2018}, pp. \bibinfo{pages}{289--305}.
\bibitem[{Yang et~al.(2022)Yang, Zhuo, Qi, Shi, and Gao}]{yang2022st++}
\bibinfo{author}{L.~Yang}, \bibinfo{author}{W.~Zhuo}, \bibinfo{author}{L.~Qi}, \bibinfo{author}{Y.~Shi}, \bibinfo{author}{Y.~Gao},
\newblock \bibinfo{title}{St++: Make self-training work better for semi-supervised semantic segmentation},
\newblock in: \bibinfo{booktitle}{Proceedings of the IEEE/CVF conference on computer vision and pattern recognition}, \bibinfo{year}{2022}, pp. \bibinfo{pages}{4268--4277}.
\bibitem[{Bengio et~al.(2009)Bengio, Louradour, Collobert, and Weston}]{bengio2009curriculum}
\bibinfo{author}{Y.~Bengio}, \bibinfo{author}{J.~Louradour}, \bibinfo{author}{R.~Collobert}, \bibinfo{author}{J.~Weston},
\newblock \bibinfo{title}{Curriculum learning},
\newblock in: \bibinfo{booktitle}{Proceedings of the 26th annual international conference on machine learning}, \bibinfo{year}{2009}, pp. \bibinfo{pages}{41--48}.

\end{thebibliography}

\end{document}